\documentclass[10pt,twocolumn,letterpaper]{article}

\usepackage{iccv}
\usepackage{times}
\usepackage{epsfig}
\usepackage{graphicx}
\usepackage{amsmath}
\usepackage{amssymb}

\usepackage{caption}
\usepackage[caption=false,font=footnotesize,labelfont=sf,textfont=sf]{subfig}
\usepackage{multirow}
\usepackage{color}
\usepackage[dvipsnames]{xcolor}
\usepackage{colortbl}
\usepackage[pagebackref=true,breaklinks=true,letterpaper=true,colorlinks,bookmarks=false]{hyperref}

\definecolor{lightskyblue}{rgb}{0.94,1.0,1.0}
\definecolor{lightgreen}{rgb}{0.94,1.0,0.94}
\definecolor{whitesmoke}{rgb}{0.92,0.92,0.92}
\definecolor{seashell}{rgb}{1.0,0.96,0.93}

\newcommand{\LX}[1]{{\color{black}#1}}
\newcommand{\LC}[1]{{\color{black}#1}}
\newcommand{\rz}[1]{{\color{black}#1}}
\newcommand{\gl}[1]{{\color{black}#1}}

\newcommand{\hf}[1]{{\color{black}#1}}
\newcommand{\HF}[1]{{\color{black}#1}}

\iccvfinalcopy 

\def\iccvPaperID{5496} 

\ificcvfinal\fi

\begin{document}

\title{3D-FRONT: 3D Furnished Rooms with layOuts and semaNTics}

\author{\textbf{Huan Fu}\textsuperscript{1} \ \ \textbf{Bowen Cai}\textsuperscript{1} \ \ \textbf{Lin Gao}\textsuperscript{2} \ \ \textbf{Lingxiao Zhang}\textsuperscript{2} \ \ \textbf{Jiaming Wang}\textsuperscript{1} \vspace{0.8mm} \\ 
\textbf{Cao Li}\textsuperscript{1} \ \ \textbf{Qixun Zeng}\textsuperscript{1} \ \ \textbf{Chengyue Sun}\textsuperscript{1} \ \ \textbf{Rongfei Jia}\textsuperscript{1} \ \ \textbf{Binqiang Zhao}\textsuperscript{1} \ \ \textbf{Hao Zhang}\textsuperscript{3}
\vspace{1.8mm}
\\
\textsuperscript{1}Alibaba Group \\
\textsuperscript{2}Institute of Computing Technology, Chinese Academy of Sciences\\
\textsuperscript{3}School of Computing Science, Simon Fraser University\\
{\tt\small \{fuhuan.fh, kevin.cbw, rongfei.jrf, binqiang.zhao\}@alibaba-inc.com} \\
{\tt\small \{gaolin, zhanglingxiao\}@ict.ac.cn}
\ \ \ \ {\tt\small haoz@sfu.ca}}

\ificcvfinal\thispagestyle{empty}\fi

\twocolumn[{%
\renewcommand\twocolumn[1][]{#1}%
\maketitle
\begin{center}
    \centering
    \includegraphics[width=0.99\textwidth]{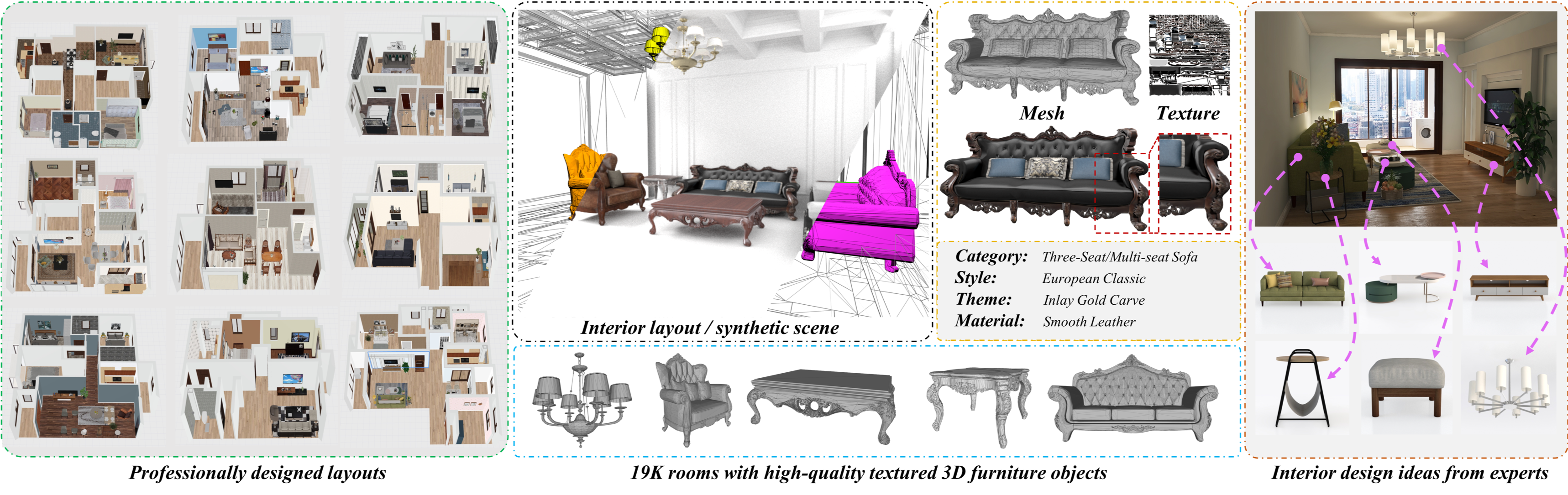} \vspace{-5pt}
    \captionof{figure}{\textbf{3D-FRONT} is a new, large-scale, and comprehensive repository of synthetic indoor scenes with professionally and distinctively designed layouts, a large number (\LX{18,968}) of rooms populated with 3D furniture objects that are stylistically compatible and endowed with high-quality textures. All freely available to the academic community and beyond.}
    \label{fig:teaser}
\end{center}%
}]

\begin{abstract}
We introduce 3D-FRONT (3D Furnished Rooms with layOuts and semaNTics), a new, large-scale, and comprehensive repository of synthetic indoor scenes highlighted by professionally designed layouts and a large number of rooms populated by high-quality textured 3D models with style compatibility. From layout semantics down to texture details of individual objects, our dataset is freely available to the academic community and beyond. Currently, 3D-FRONT contains \LX{18,968} 
rooms diversely furnished by 3D objects, far surpassing all publicly available scene datasets. In addition, the \LX{13,151} furniture objects all come with high-quality textures. 
While the floorplans and layout designs are directly sourced from professional creations, the interior designs in terms of furniture styles, color, and textures have been carefully curated based on a recommender system we develop to attain consistent styles as expert designs. Furthermore, we release Trescope, a light-weight rendering tool, to support benchmark rendering of 2D images and annotations from 3D-FRONT. 
We demonstrate two applications, interior scene synthesis and texture synthesis, that are especially tailored to the strengths of our new dataset.

\end{abstract}

\setlength\tabcolsep{7pt}
\begin{table*}[t!]
  \centering
    \begin{tabular}{c|c|r|r|c|c}
    \hline
    \textbf{Dataset} & \textbf{Layout Design} & \textbf{\#3DFRs}  & \textbf{\#CAD models} &\textbf{Model Textures} & \textbf{3D Annotation} \\
    \hline\hline
    NYU-Depth v2 \cite{Silberman:ECCV12} & Real scan  & N/A  & N/A & No texture & Raw RGB-D \\
    TUM \cite{sturm2012benchmark}  & Real scan  & N/A & N/A  & No texture & Raw RGB-D  \\
    \hline
    SUN3D \cite{xiao2013sun3d} &  Real scan  & 254 & N/A & No texture  & Raw PCD \\
    BuldingParser \cite{armeni20163d} & Real scan  & 270 & N/A & No texture   & Raw PCD \\
    SceneNN \cite{hua2016scenenn} & Real scan  & 100  &  N/R & Rec.~from Scan  & Raw Mesh  \\
    Matterport3D \cite{chang2017matterport3d} & Real scan  & 2,056   & N/A  & Rec. from Scan & Raw Mesh  \\
    ScanNet \cite{dai2017scannet} & Real scan  & 1,506   & 296 & Rec.~from Scan &  Raw Mesh  \\
    Scan2CAD \cite{avetisyan2019scan2cad} & Real scan & 1,506  & 3,049 & No texture & Mesh      \\
    OpenRooms \cite{li2020openrooms} & Real scan & 1,068  & 2,500 & Amateur & Mesh      \\
    \hline
    SceneNet \cite{handa2016understanding} & \hf{Professional} & 57   & N/R  & No texture & Mesh   \\
    InteriorNet \cite{li2018interiornet} & Professional & N/A & N/A & No texture & N/A \\
    Hypersim \cite{roberts2020hypersim} & Professional & N/A  & N/A & Per-pixel color & RGB-D      \\    
    Structured3D \cite{zheng2019structured3d} & Professional &  N/A  & N/A &  No texture  & 3D structures  \\
    \hline
    3D-FRONT  & Professional & {\bf \LX{18,968}}  & {\bf \LX{13,151}} & {\bf Professional} & Mesh   \\
    \hline
    \end{tabular}%
  \caption{\textbf{\LC{Comparison} between prominent 3D indoor scene datasets}, 
where ``\#3DFRs" represents the number of rooms or scenes populated with 3D furniture objects, 
``N/A" = ``not available", ``N/R" = ``not reported", ``Raw Mesh" \LC{denotes} machine reconstructed meshes, and ``Raw PCD'' refers to reconstructed point clouds. For model textures, ``Rec.~from Scan" is the result of reconstruction from raw RGB-D data, while ``Amateur'' and ``Professional'' refer to who designed the textures. The ``3D structures'' annotatd by Structured3D~\cite{zheng2019structured3d} contain information on primitives including~\gl{3D boxes} and their relations.}
\vspace{-2mm}
  \label{tab:statistics}%
\end{table*}%

\section{Introduction}
\label{sec:intro}

The computer vision community has invested much effort into the study of 3D indoor scenes, from 3D reconstruction, visual SLAM, and navigation, to scene understanding, affordance analysis, and generative modeling. With data-driven and learning-based approaches receiving more and more attention in recent years, there has been a steady accumulation of indoor scene datasets~\cite{Silberman:ECCV12,sturm2012benchmark,xiao2013sun3d,armeni20163d,hua2016scenenn,chang2017matterport3d,dai2017scannet,handa2016understanding,li2018interiornet,zheng2019structured3d,li2020openrooms} to drive the deep learning revolution that has redefined the landscape of indoor scene processing. 




Existing 3D scene datasets all fall into two broadly categories: acquired (via scanning and reconstruction) vs.~designed (i.e., synthetic scenes created by humans). In terms of data volume, the largest repository is ScanNet~\cite{dai2017scannet} which consists of 2.5M RGB-D images from 1,513 scanned real scenes acquired by commodity sensors, in 707 distinct spaces. The 3D scenes, including textured 3D objects, were recovered by state-of-the-art 3D reconstruction techniques from the raw scans, which are typically noisy and incomplete. As a result, the reconstructed meshes are often of low quality, both in geometric fidelity and texture quality.


In the world of synthetic 3D indoor scene datasets, the recent exit by SUNCG~\cite{song2016ssc} has left an apparent void in the community.
Most recently, Structured3D~\cite{zheng2019structured3d} and OpenRoom~\cite{li2020openrooms} have emerged as promising alternatives. In addition to providing professionally designed room layouts, Structured3D~\cite{zheng2019structured3d} aims to provide \LC{large-scale} photo-realistic scene {\em images\/} with rich 3D structure annotations. However, the actual 3D furniture objects populating the scenes \LC{are} not included in the dataset. OpenRoom~\cite{li2020openrooms} replaces detected objects in a set of 1,068 scanned scenes from ScanNet~\cite{dai2017scannet} with CAD models from ShapeNet~\cite{chang2015shapenet}. A major contribution of this dataset is to provide ground-truth annotations of complex material parameters for the CAD objects. However, the dataset has not been released at this point and according to the authors' account, only 2.5K CAD models were annotated with material properties. 


In this paper, we introduce 3D-FRONT (3D Furnished Rooms with layOuts and semaNTics), a new, large-scale, and {\em comprehensive\/} repository of synthetic 3D indoor scenes\LC{. It contains} professionally and distinctively designed layouts spanning \LX{31} scene categories, object semantics (e.g., category, style, and material labels), and a large number (\LX{18,968}) of rooms populated with 3D furniture objects. Most importantly, these 3D furniture objects are all endowed with high-quality textures, thanks to 3D-FUTURE~\cite{fu20203dfuture}, a recently released dataset of quality 3D furniture used in industrial productions. Furthermore, the selection of furniture objects from 3D-FUTURE to populate the scenes in 3D-FRONT has been inpired by expert interior designs. Specifically, the selection is based on a {\em recommender system\/} learned from the expert designs, while taking into account of furniture styles both in terms of geometry and texture. As a result, the furnished rooms in 3D-FRONT consist of {\em stylistically compatible\/} objects adhering to the design inspirations.

In Table~\ref{tab:statistics}, we present essential information for the current public release of 3D-FRONT and compare to other prominent indoor scene datasets. As we can see, the most compelling feature of our dataset is the large number of 3D furnished rooms, which far surpasses all the other publicly available datasets. Style compatibility, as well as the high texture quality, of the furniture objects in each scene (see middle of Figure~\ref{fig:teaser}) is another unique attribute of 3D-FRONT. On top of all these, the total number of rooms with professionally designed layouts is much larger than \LX{18,968}; it is close to 45,000. 
%
%
Last but not least, we share Trescope, a light-weight rendering tool, with the community so that the users of 3D-FRONT can easily capture their desired 2D renderings and annotations to guide their image-driven learning tasks. 
We will continuously improve 3D-FRONT by releasing an industrial rendering engine (AceRay) and providing much enriched texture and 3D geometry contents.

We anticipate that 3D-FRONT, being as comprehensive as it is, will enable and further drive a whole suite of AI-powered and data-driven scene analysis and modeling applications. We demonstrate two applications which cannot be well supported by other publicly available datasets --- these applications are best served by having a large number of high-quality textured mesh models with style consistency, a unique feature of 3D-FRONT. One such application is learning to texture 3D objects in indoor scenes. In another, by learning the layout of 3D furniture 
in each room with \cite{wang2018deep}, we can coherently predict and arrange functional furniture for an empty room.


\section{Related Work}
\label{sec:related}

Over the past years, a large number of RGB-D benchmarks have been constructed and made publicly available \cite{koppula2011semantic,armeni2017joint,Silberman:ECCV12,sturm2012benchmark,xiao2013sun3d,song2015sun,hua2016scenenn,chang2017matterport3d,dai2017scannet,handa2016understanding,mccormac2017scenenet,li2018interiornet,zheng2019structured3d,li2020openrooms,armeni20163d,savva2016pigraphs,straub2019replica,xia2018gibson,fisher2012example,garcia2018robotrix,zhang2017physically}. Current 3D scene datasets are mainly collected based on \gl{scanning and reconstruction or human creation}. 
These datasets thus fall into two broadly categories: Acquired vs.~Designed.

\vspace{5pt}

\noindent \textbf{Acquired Scenes.} To construct ``Acquired" datasets, researchers capture RGB-D videos, reconstruct the scene meshes, and manually label the frames or the reconstructed scenes. For example, NYU-Depth v2 \cite{Silberman:ECCV12} gathered 464 short RGB-D sequences from different rooms via Kinect, where 1,449 images are selected and labeled with pixel-level annotations. 
SUN RGB-D \cite{song2015sun} collected 10,335 RGB-D images and provided more 2.5D annotations, such as 2D polygons and 3D bounding boxes correspondences, room layouts, and scene categories. These datasets may lack the physical relationship between the frames and the scene space's real 3D structure. To address the issue, SUN3D \cite{xiao2013sun3d} developed an interactive reconstruction pipeline to recover the 3D scene structures for 254 different spaces in 41 buildings, in which 8 scenes are provided with semantic labels for 3D point clouds and camera poses. SceneNN \cite{hua2016scenenn} improved the pipeline by recovering mesh surfaces instead of point clouds for 100 scans. Further, one of the largest ``Scanned" datasets, \emph{i.e.,} ScanNet \cite{dai2017scannet}, has been established. It reconstructed 1,513 rooms based on 2.5M RGB-D views, and labeled rich 3D annotations, including estimated 3D camera poses, surface reconstructions, semantic segmentation, and 2D-3D alignments. 

Since 3D scene reconstruction \gl{with fine geometric and textures details} is still a \gl{challenging} problem with the \gl{depth cameras on the shelf such as Kinect}, \gl{the mesh qualities in these scene dataset are usually not as good as the synthetic data.} Besides, some of the 3D annotations may be unreliable or imprecise \gl{due to the reconstruction error}, such as camera pose and 2D-3D alignment. 

\vspace{5pt}

\noindent \textbf{Designed Scenes.} 
\gl{Another type of scene dataset is from the human creation with professional design software as this 3D-FRONT.} \gl{In addition to 3D-FRONT, there is one synthetic (designed) dataset that shares both the layout and the well-posited 3D CAD mesh models, \emph{i.e.,} SceneNet \cite{handa2016understanding} with providing 59 scenes.} Several other synthetic benchmarks share 2D and 2.5D contents based on designed synthetic scenes. For example, InteriorNet \cite{li2018interiornet} released 15k sequences and 5M images, which are rendered from their large-scale scene packages. Further, Structure3D \cite{zheng2019structured3d} provided 21,835 panoramic images with the corresponding structure annotations, such as panoramic layouts, depth, surface normal. Recently, Li et al.~\cite{li2020openrooms} built OpenRooms, a synthetic benchmark based on ScanNet, and planned to share rendered images with their high-quality SVBRDF and spatially-varying lighting. 
Also, \rz{Hypersim~\cite{roberts2020hypersim} presented a photorealistic synthetic dataset for holistic indoor scene understanding, focusing on providing per-pixel depth and disentangled illuminance and reflectance properties over scene images designed by professional artists.}

These large-scale synthetic datasets have not made the \gl{completed }scene packages, including the floorplans' mesh, \gl{the large amount of involved CAD models with fine geometric and texture details, and the layout with }design ideas, publicly available. In contrast, 3D-FRONT shares everything that is used to construct houses, from real layouts to interior design ideas and involved objects. The holiest repository of indoor scene packages enables a robot to navigate in them. It also allows the researchers to render whatever information they need for new subjects studying.


\begin{figure*}[t!]
    \centering
        \includegraphics[width = 0.98\linewidth]{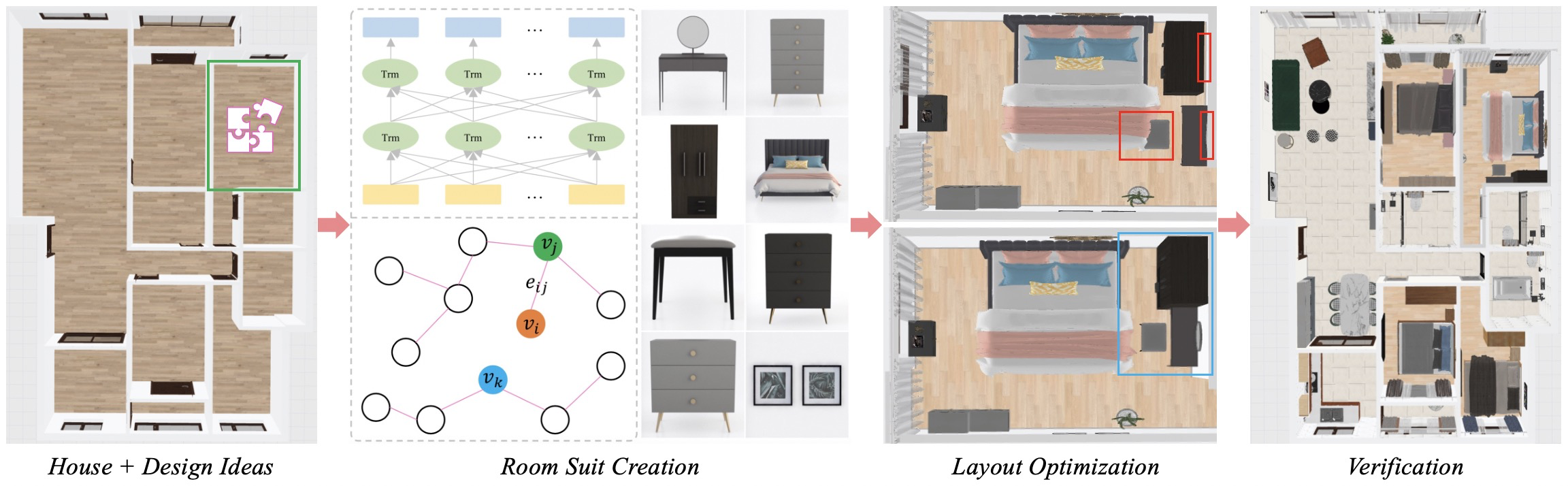}
    \caption{\LC{\textbf{Pipeline of building 3D-FRONT.}}
    We start from an empty house with professional design ideas, create the room suites, optimize the layouts \rz{(e.g., to resolve artifacts highlighed in the red boxes)}, and finally verify the furnished rooms.} 
    \vspace{-2mm}
    \label{fig:pipeline}
\end{figure*}

\begin{figure}[t!]
    \centering
        \includegraphics[width = 0.98\linewidth]{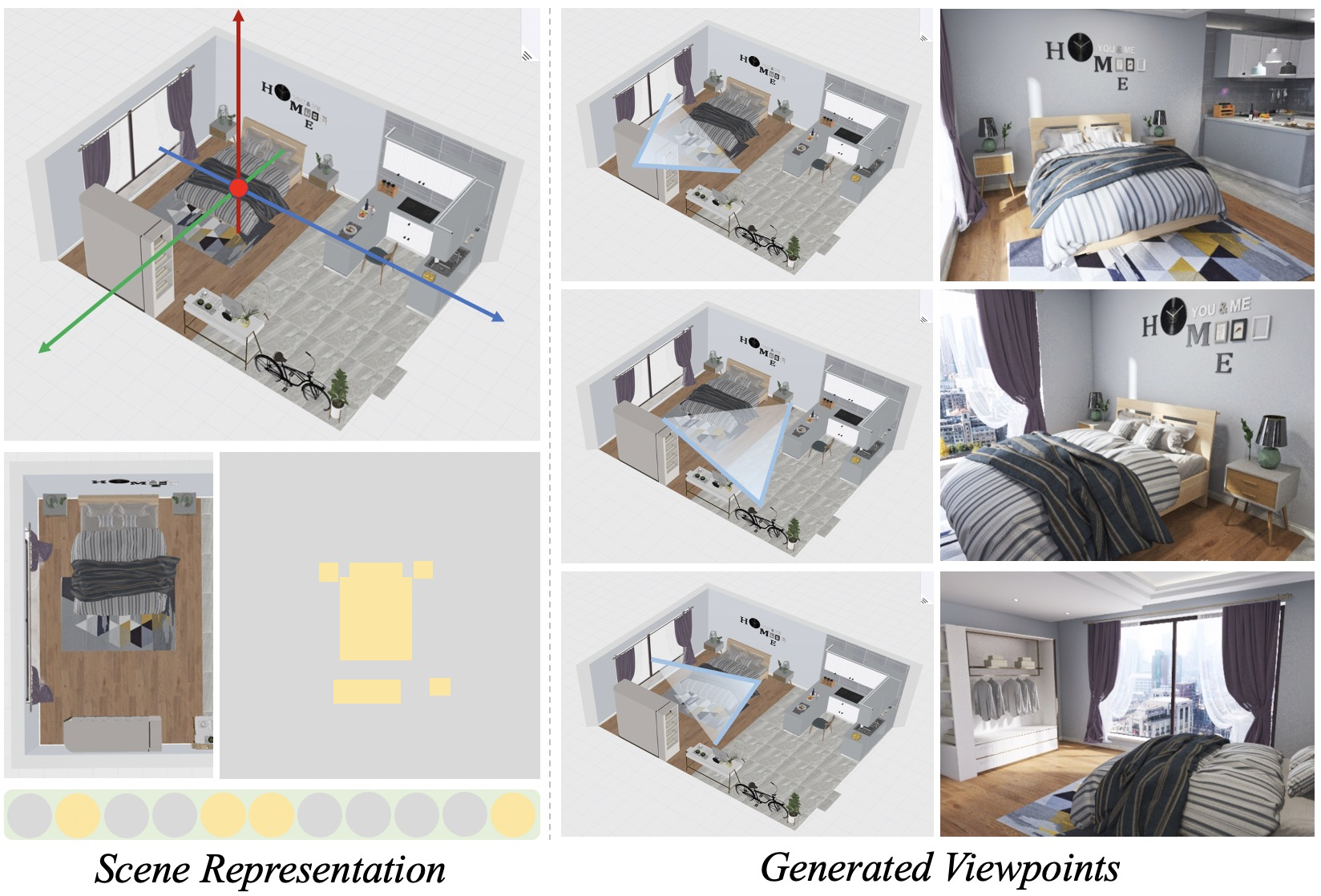}
    \caption{\textbf{Viewpoint Generation.} Each scene is associated with several natural camera views to facilitate rendering.}
    \vspace{-5mm}
    \label{fig:viewpoints}
\end{figure}

\section{Building 3D-FRONT}
\label{sec:building_3dfront}

Creating a large-scale 3D scene repository such as 3D-FRONT is a non-trivial task. Our 3D-FRONT project has been built on a large volume (about 60K) of professionally designed houses and 1M 3D CAD \rz{meshes. While we are unable to publish all these meshes, due to copyright restrictions, all the models and learning algorithms employed during the data collection progress have been trained on the large database.} As shown in Figures~\ref{fig:pipeline} and~\ref{fig:viewpoints}, we start from some house collections, create room suites, optimize the layout, verify the created interior designs, and finally assign qualified camera viewpoints. In the following, we will detail the pipeline as well as the techniques involved. 

\subsection{Room Suite Creation}

Given a synthetic house and its professional design ideas, we automatically create room suites for the scenes. Here, the design ideas for a room consist of \rz{the category labels of the objects, and their positions, orientations, sizes, and styles.} Taking a bedroom as an example, we first randomly select a seed object, \emph{e.g.,} a bed, from a 3D model pool according to the required size and style. We then recurrently identify the visually matched furniture according to the room suite thus far \rz{until the room is filled.}

We mainly rely on the Furnishing Suite Composition (FSC) approach in 3D-FUTURE \cite{fu20203dfuture} to create visually compatible suites. Specifically, leveraging on the large-scale expert scene designs, we \rz{carry out} two tasks, \emph{i.e.,} mask prediction and suite compatibility scoring, to model visual compatibility. The first task predicts the masked (removed) furniture given other objects in a suite. And the second task evaluates the compatibility score of the input suite. We utilize a textured image to represent each object (furniture). The two tasks optimize a visual embedding network (VEN) \cite{sandler2018mobilenetv2} and two transformer architectures \cite{vaswani2017attention,devlin2018bert}, so that the trained VEN can extract informative visual feature for each object. With the learned visual representation and the given attributes, including category, style, color, material, and size, for each object. We train gradient boosting decision trees (GBDT) \cite{friedman2001greedy} to infer decision rules based on these information, and post a logistic regression (LR) layer to estimate the comparability scores of the room suites. \gl{These two techniques are integrated as the GBDT-LR model.}

3D-FUTURE \cite{fu20203dfuture} first adopts the visual embedding extracted from VEN to perform a primary ranking, then employs the trained GBDT-LR model to re-rank the selected candidates for online recommendation. We improve the primary ranking stage by considering graph auto-encoder techniques \cite{cucurull2019context}. In detail, we define an undirected graph $\mathcal{G} = \{\mathcal{V}, \mathcal{E}\}$, and learn a graph auto-encoder (GAE) for visual compatibility prediction following \cite{cucurull2019context}. The graph nodes are all the involved objects in the designed house database. Each node is represented with a feature vector extracted from VEN. Each edge's weight is equal to $1$ if the two objects are visually appealing, and 0 otherwise. With the graph, we first learn a graph convolutional network (GCN) \cite{kipf2016semi} as an encoder to propagate neighborhood information to obtain new representations, depending on the connections. Then, we adopt a fully connected layer as a decoder to reconstruct the weight matrix. When building 3D-FRONT, we use the trained models to perform recommendation from the 3D-FUTURE pool \cite{fu20203dfuture}.


\subsection{Layout Optimization and Verification}

\rz{We observed that with the room suites constructed using the techniques described so far, placing objects into the corresponding rooms according to their suggested positions and orientations, various layout artifacts still remained.} For example, a bed may overlap with its nearby nightstand in the 3D space. Other examples are highlighted in red boxes in Figure~\ref{fig:pipeline}. One of the main reasons is that it is difficult to find a visually matched furniture based on concurrent room suite that, at the same time, has the same size required by the design ideas --- \rz{there is potential conflict between style and size compatibilities}. To this end, we apply the layout optimization algorithm proposed in \cite{weiss2018fast}. 

Specifically, we start from the initially created designs, and slightly modify the \rz{object positions in the room suites} in order to satisfy several layout constraints in \cite{weiss2018fast}, including pairwise distance, focal point distance, distance to wall, accessibility, and collision. \rz{These constraints were constructed based on polling statistics of the design rules from our synthetic house database. Since the intial layouts often provide a good starting point, we only optimize the defined energy function in up to 50 iterations. On average, the optimization only takes 10s for each room.}

We further verify the created designs and remove the unsatisfied ones to ensure dataset quality. To facilitate the reviewing step, we develop a light-weight renderer Trescope that enables the reviewers to browse the synthetic houses online in an interactive manner. Note that, Trescope supports offline benchmark rendering on local machines for 3D-FRONT. The renderer will be shared so that users of 3D-FRONT can capture their desired renderings such as images, depth, normal, and segmentation. 

\subsection{Viewpoint Generation}
The viewpoint generation stage aims to assign several cameras to each scene, and ensure most of the cameras have practical viewpoints. For the purpose, we recast this problem as a similarity measure problem, thus mainly transfer knowledge from expert ideas. Specifically, we are provided with many excellent scenes with suggested camera viewpoints by expert designers (about 5,000 rooms in our database). Given a scene, we choose a ``center" object as the origin to build the world coordinate. We compute the normalized distances to define an object group (like a graph), and convert its mask projection as a feature vector to represent the scene structure. With this method, we can compute cosine similarity to perform scene retrieval, thus generate practical camera viewpoints for new scenes. Note that, for the created scenes, we are allowed to extract different scene features by selecting different ``center" objects. This would guarantee the diversity of the generated viewpoints.


\begin{figure*}[!t]
\centering
    \includegraphics[width=\textwidth]{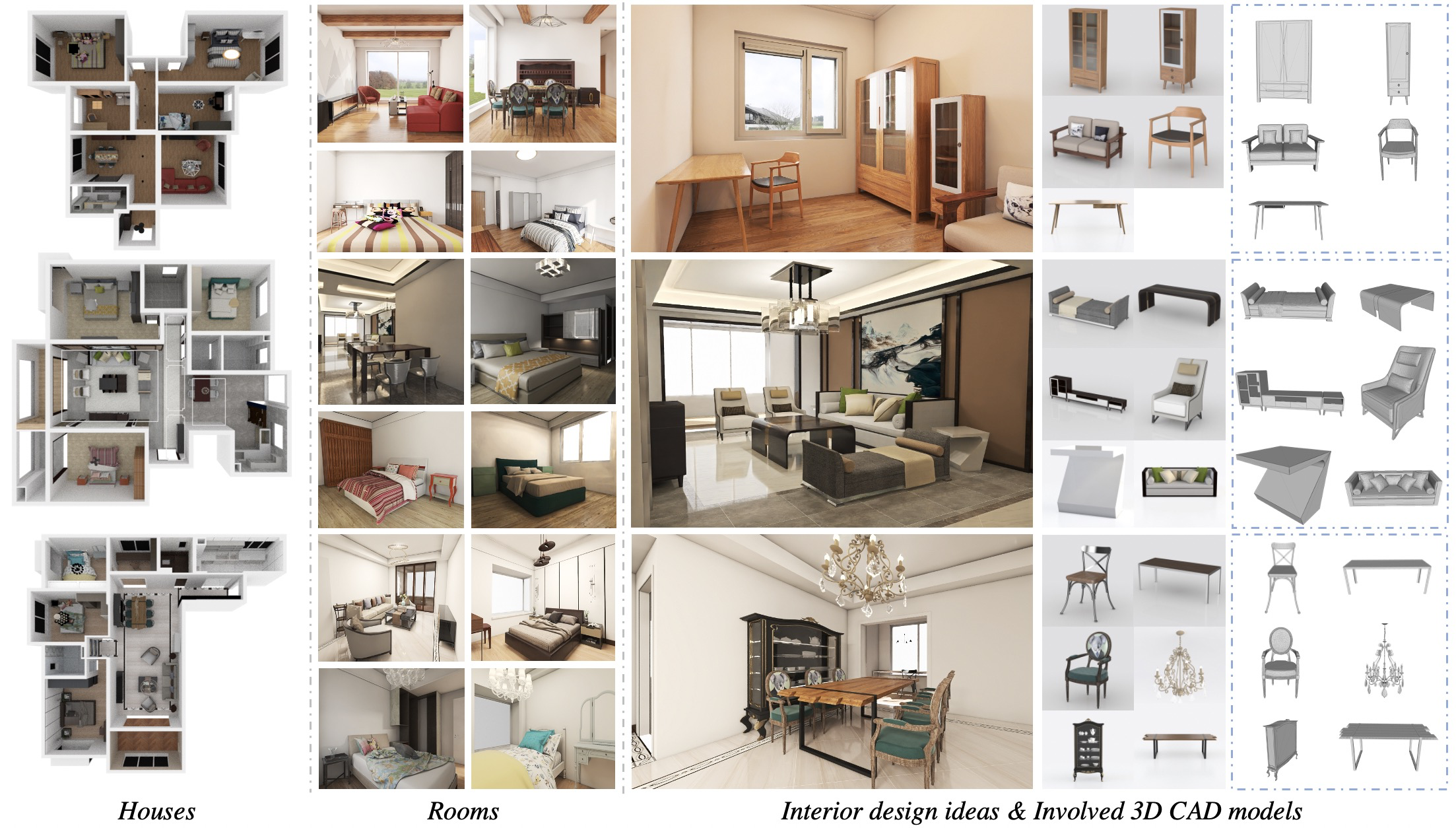}
   \caption{\textbf{House Examples in 3D-FRONT.} The left column shows the top-down views of three houses. The middle column presents several rooms contained in these houses, including bedrooms, living rooms, dining rooms, \emph{etc}. The interior design ideas at the right column summarize the textured objects involved in the rooms and their high-quality 3D CAD models.}
\vspace{-4mm}
\label{fig:drinet}
\end{figure*}

\section{Validation and Assessment}
\label{sec:vcp}

\rz{In this section, we offer several means to validate and assess the way our dataset was built and the quality and utility of the data. Applications are discussed in Section~\ref{sec:apps}.}


\vspace{-8pt}

\paragraph{Evaluation of recommender system.}
We collected 8,000 scene designs and their design logs from the online Topping Homestyler platform\footnote{https://www.homestyler.com} for our evaluation. We discuss several metrics, including Area Under The Curve (AUC)~\cite{fogarty2005case}, 1-N Average Rank (1-N Avg Rank), N-1 Average Rank (N-1 Avg Rank), 1-N Hit@10, and 1-N Hit@20, to show the superiority of the designs in 3D-FRONT. These metrics are calculated based on experts' online logs.
To explain these measurements, we take a room \LX{suite} (Bed, Nightstand, Chair) $\Leftrightarrow$ (A, B, C) as an example, where a designer chooses the objects A, B, and C in order. 1-N Avg Rank means that we recurrently perform (A) $\to$ Nightstand and (A, B) $\to$ Chair, respectively, and compute the average rank (B and C). Here, Nightstand and Chair are the required categories, and B and C are the specific objects. N-1 denotes that we recommend each object given the other two. Hit@K calculates the TopK recall accuracy. For (A, B) $\to$ Chair, a correct recommendation in TopK means that C ranks less than K. We refer to the supplementary materials for more details about these metrics.


\setlength\tabcolsep{9pt}
\begin{table}[t!]
\centering
\begin{tabular}{ c | c || c | c  }
\hline
\multicolumn{2}{c||}{Metrics}  & FSC \cite{fu20203dfuture} & FSC + GAE \cite{cucurull2019context} \\
 \hline
 \multirow{3}{*}{$\uparrow$}
 & AUC & 0.766 & \textbf{0.772} \\
 & 1-N Hit@10 & 33.6\% & \textbf{36.1\%} \\
 & 1-N Hit@20 & 61.3\% & \textbf{64.3\%} \\
 \hline
 \multirow{2}{*}{$\downarrow$}
 & 1-N Avg Rank & 41.6 & \textbf{37.3} \\
 & N-1 Avg Rank & 26.7 & \textbf{24.1} \\
 \hline
\end{tabular}
\caption{\textbf{Evaluating the Pipeline.} $\uparrow$: higher is better. $\downarrow$: lower is better. We perform recommendation based on a extremely large 3D pool (about 1M models). When calculating these scores, invalid items in the retrieval sequences have been filtered out based on fine-grained category labels.}
\label{tab:vcp}
\end{table}

\begin{table}[t!]
\centering
\begin{tabular}{ c | c | c }
\hline
 & Questions & 3D-FRONT \\
\hline\hline
\multirow{4}{*}{Scene}
& Plausible Layout & \textbf{62.5\%} \\
& Design Quality & \textbf{69.2\%} \\
& Richer Texture & \textbf{70.0\%} \\
& Style Compatibility & \textbf{65.4\%} \\
\hline\hline
\multirow{2}{*}{3D Model}
& Richer Texture & \textbf{65.4\%} \\
& Preferable & \textbf{61.5\%} \\
\hline
\end{tabular}
\caption{\rz{\textbf{User studies on data quality: 3D-FRONT \emph{vs.}~SUNCG.} The reported percentages indicate how many users on AMT chose scenes/models from 3D-FRONT when presented questions regarding the quality criteria.}}
\vspace{-5mm}
\label{tab:dataset-user-study}
\end{table}


The qualitative scores are reported in Table~\ref{tab:vcp}. Generally, incorporating GAE \cite{cucurull2019context} with the original FSC \cite{fu20203dfuture} would yield improvements on all metrics. We point out that both FSC and its improved version (FSC+GAE) can generate high-quality room \LX{suites}, though it seems that the performance numbers of 1-N Hit@10 (33.6\%$\sim$36.1\%) and 1-N Avg Rank (41.6$\sim$37.3) are not significant. But it should note that our 3D pool contains more than 1M models. The vast collection makes the visual compatibility inspired recommendation task extremely challenging, though we have filtered out invalid items in the retrieval sequences according to the fine-grained category labels. \LC{It's} also worth to mention that, after layout optimization and verification, our AI created designs (room \LX{suites} + professional design ideas) have been used for VR shopping by eCommerce merchants. The rate of our high-quality designs (or customer preferred designs) is 88\%, while it is only 71\% for designs from ordinal-level designers. The comparison may not fair for ordinal-level designers since we reuse professional design ideas. However, it strongly supports that the shared scene designs in 3D-FRONT are really with high-quality. 

\vspace{-8pt}

\paragraph{User study.}
\rz{We conduct a series of user studies, on Amazon Mechanical Turk (AMT), to assess the quality of the data provided by 3D-FRONT, in comparison with SUNCG \cite{song2016ssc}. The quality criteria considered include those related to scene layouts (in terms of plausibility, design quality, and richness of texture) and individual objects (in terms of texture quality and preferability), as well as style compatibility.
We refer to the supplemental material for more details on each study.
As for the user study setting, we randomly sampled 90 pairs of scenes and 30 pairs of 3D models from 3D-FRONT and SUNCG based on scene type and model category. Each pair was labeled by 20 master-level annotators in AMT. Thus, the scene and model scores are calculated using 1,800 and 600 feedback, respectively.  

From the scores reported in Table~\ref{tab:dataset-user-study}, we see that for {\em each\/} quality criterion assessed, the majority of Turkers (between $60\%$ and $70\%$), preferred data presented by 3D-FRONT. We believe that higher-quality datasets would not only lead to improved performance of algorithms which are trained on these datasets, but also enable new applications. It should also be evident that most, if not all, applications in the computer vision and graphics community which had utilized SUNCG, would also be well supported by 3D-FRONT.}

\vspace{-8pt}



\begin{figure}[t]
			\centering
			\includegraphics[width = \linewidth]{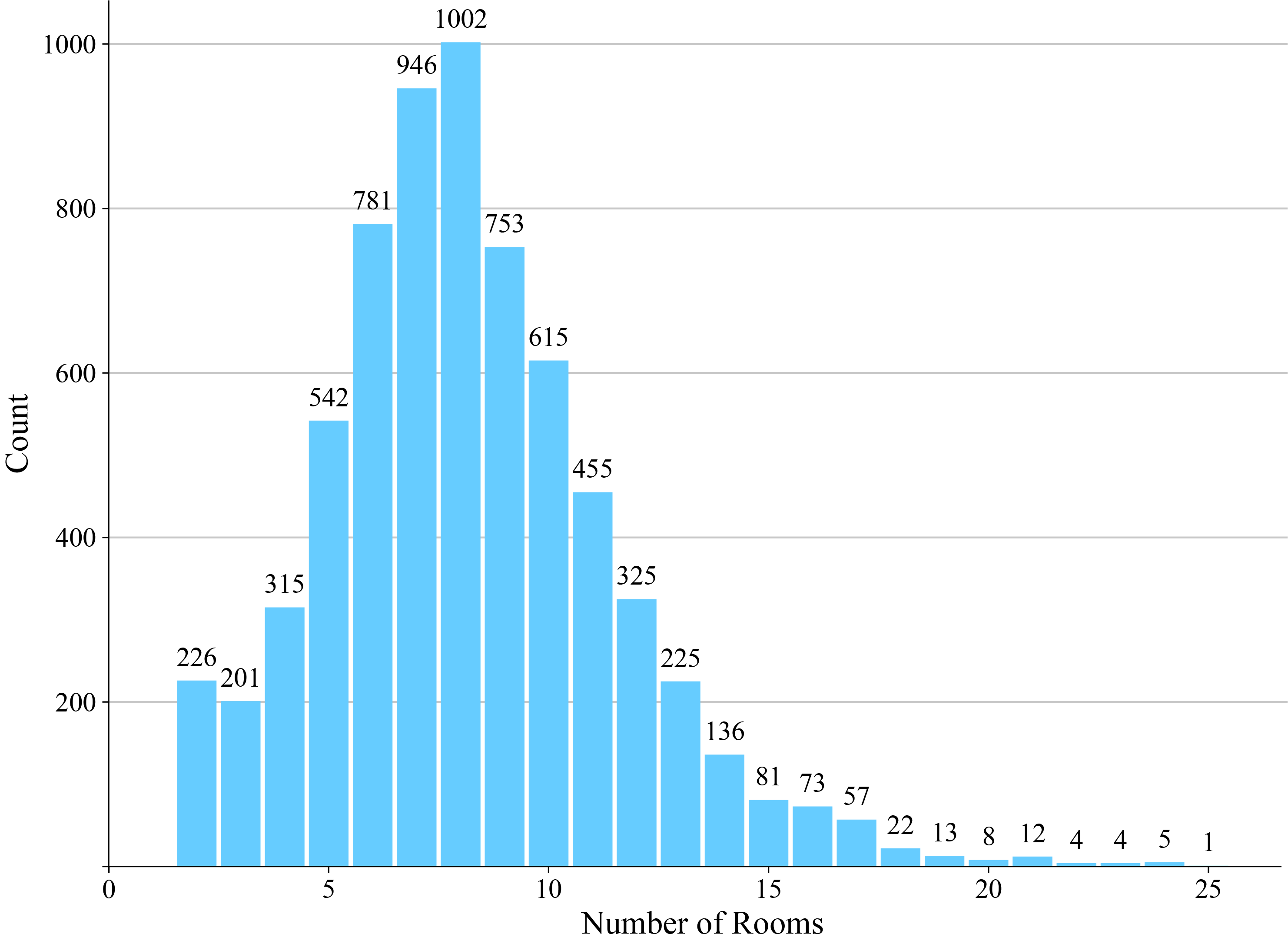}
			\caption{
			\gl{\textbf{Statistics of room numbers \LC{per} house}. 3D-FRONT contains 6,813 distinct houses constructed by \LX{44,427} rooms. There are 6.5 rooms \LC{per house on average}.}  
			}
			\vspace{-4mm}
			\label{fig:room_num}
\end{figure}

\begin{figure}[t]
			\centering
			\includegraphics[width = \linewidth]{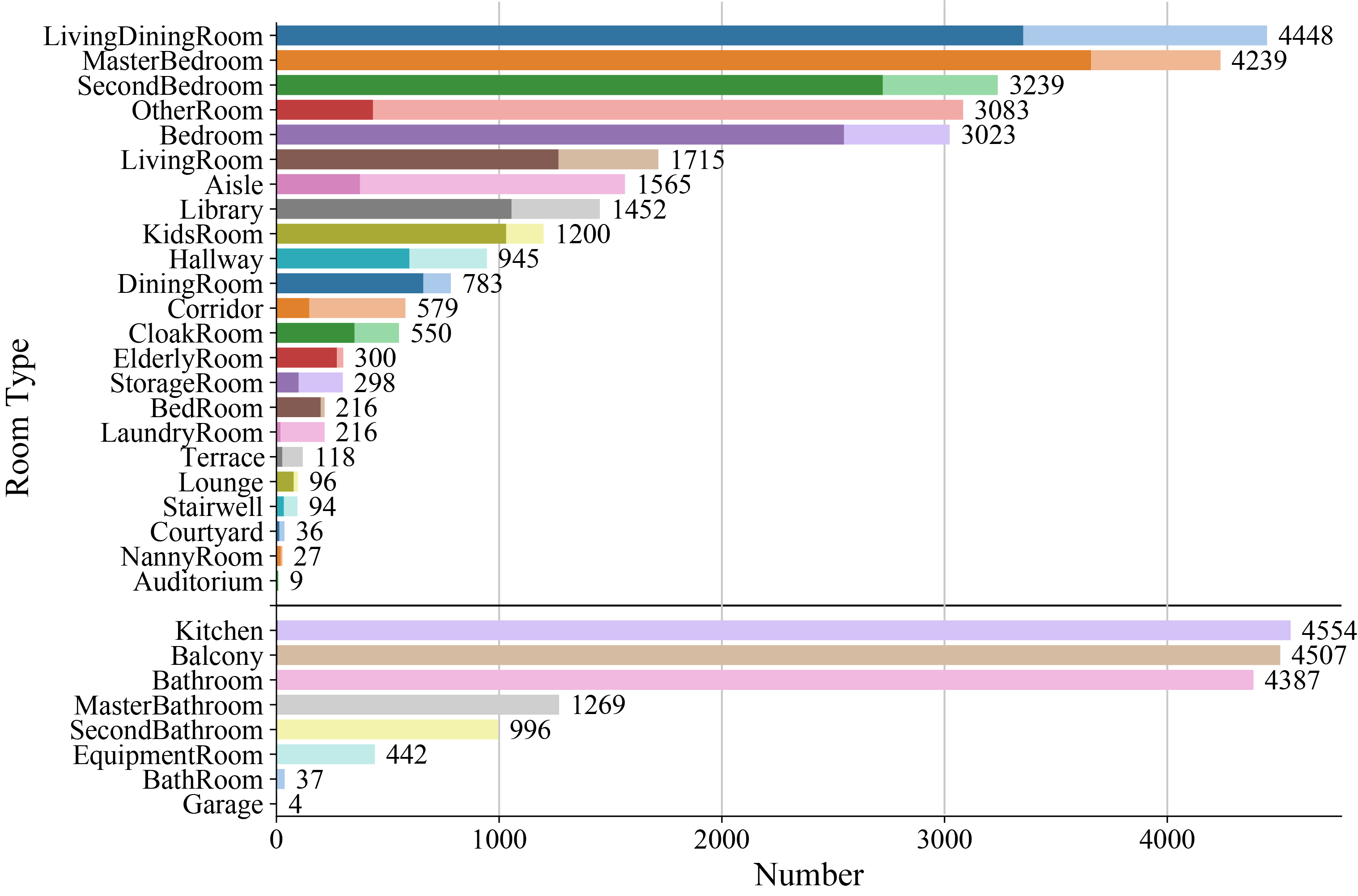}
			\caption{\textbf{Distribution of the \rz{room scenes} available in 3D-FRONT, organized by type.} There are \LX{44,427} rooms in total. A large percentage of rooms (indicated by dark color) in the top part are diversely furnished (\LX{18,968}). \LC{These} rooms, such as bedrooms, living rooms, dinging rooms, and study rooms, are the activity spaces where people tend to spend most of their times living indoors.}
			\vspace{-4mm}
			\label{fig:room_type_num}
\end{figure}

\rz{
\paragraph{Properties of 3D-FRONT.}
One of the most desirable features of our dataset is that it {\em publically} shares all the essential data that would enable the modeling of {\em high-quality} indoor scene, from layout semantics down to stylistic and texture details of individual objects. While the layout ideas are directly sourced from professional designs, the interior designs are transferred from expert creations followed by a post verification process. Figure~\ref{fig:drinet} shows some additional house examples from our dataset. 

3D-FRONT enables a variety of AI-powered tasks related to 3D scenes, including data-driven designing studies, such as floorplan synthesis, interior scene synthesis, and scene suites compatibility prediction, that other scene datasets do not support adequately. It also benefits the study of 3D scene understanding subjects, such as SLAM, 3D scene reconstruction, and 3D scene segmentation. 

Figures~\ref{fig:room_num} and~\ref{fig:room_type_num} reveal some relevant statistics related to our dataset, with more that can be found
in the supplemental material. Further, we assign selected practical camera viewpoints to furnished the scenes and release Trescope, a light-weight rendering tool compatible with 3D-FRONT. These would allow users of 3D-FRONT to easily render images and annotations to support their 2D vision studies. Last but not least, we will continuously improve 3D-FRONT by adding more features. A certain plan is to share much enriched texture and 3D geometry contents.
}

\section{Applications}
\label{sec:apps}


\rz{We present two applications, interior scene synthesis and object texturing in scene contexts, to demonstrate the utility of our dataset. This only represents a small sampler of applications that can benefit from 3D-FRONT.}


\subsection{Interior Scene Synthesis}


The main goal of current scene synthesis methods \cite{wang2018deep,ritchie2019fast,li2019grains,wang2019planit,zhang2020deep} is to coherently predict and arrange functional furniture shapes. The extensive professional layout designs provided by 3D-FRONT may be immensely valuable to support the development of learning-based methods for this synthesis task. 

\rz{
Our demonstration uses the state-of-the-art neural scene synthesis method of Wang et al. \cite{wang2018deep}, where each 3D scene is represented in an orthographic top-down view, which constitutes depth, room mask, wall mask, object mask, and orientation. Their method 
trains a deep convolutional neural network to iteratively capture scene priors, so as to decide whether to add a next object, what category of object to add and where, and finally insert an instance of that object category with estimated rotation into the scene.} Following \cite{wang2018deep}, we conduct our experiment on two scene types, \emph{i.e.,} bedroom (Bedroom, MasterBedRoom, and SecondBedRoom) and living room (LivingRoom and LivingDiningRoom), and remove the rooms whose width or length is larger than 6 meters. As a result, we obtain 6,230 bedrooms and 645 living rooms, with 6,070 / 485 rooms for training and 160 / 160 rooms for evaluation. We refer to \cite{wang2018deep} for more details on training and test settings.

\begin{figure}
    \centering
        \includegraphics[width=\linewidth]{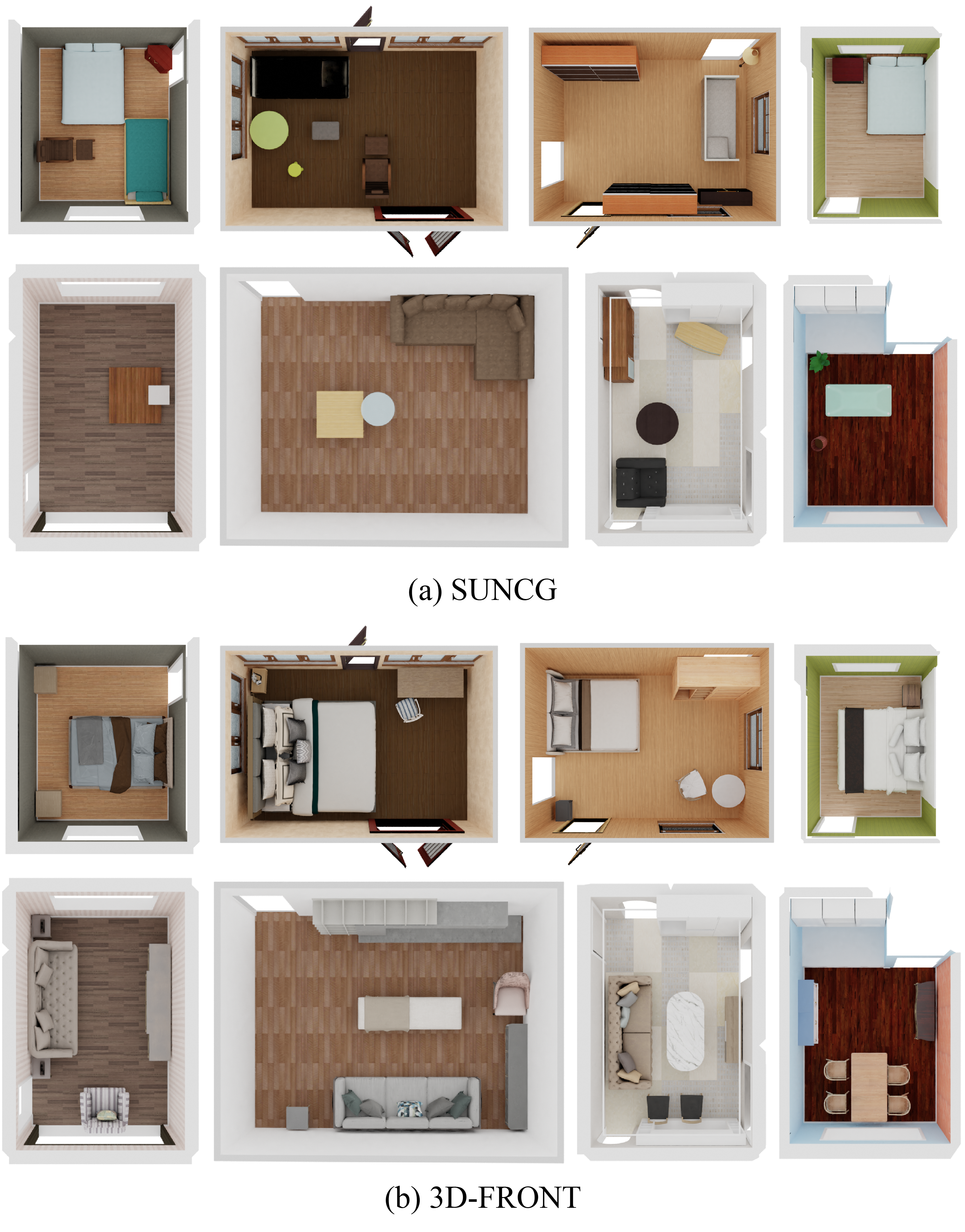}
    \caption{\textbf{Interior Scene Synthesis.} Several scenes produced by a state-of-the-art network trained on SUNCG (a) and 3D-FRONT (b), respectively. \rz{The results were synthesized from {\em randomly\/} chosen empty rooms. In each set, the first row is for bedrooms and the second row for living rooms. The 3D-FRONT results tend to show a richer variety of objects and more plausible scene layouts.}}
    \vspace{-4mm}
    \label{fig:scene_gen}
\end{figure}



\rz{We evaluate diversity of the synthesized results using converge (COV) and minimum matching distance (MMD)~\cite{achlioptas2018learning} measured by Chamfer Distance (CD) or Earth-Mover Distance (EMD)} between scenes synthesized by models trained on 3D-FRONT and on SUNCG, respectively. The results were generated from empty rooms in the combined test set of 3D-FRONT and SUNCG. For each synthesized scene, we randomly sample 100K points and calculate these metrics against the ground truth. \rz{Recall that lower MMD and higher COV indicate better synthesis ability of a method. Quantitative comparisons in Table~\ref{tab:diversity} show the dataset advantage of 3D-FRONT over SUNCG.
A qualitative comparison is shown in Figure~\ref{fig:scene_gen}.}

\setlength\tabcolsep{6pt}
\begin{table}[!t]
    \centering
     \begin{tabular}{c | cccc}
     \hline
          & MMD- & MMD- & COV- & COV- \\
          & CD $\downarrow$ & EMD $\downarrow$ & CD $\uparrow$ & EMD $\uparrow$ \\
          \hline\hline
    SUNCG \cite{song2016ssc} &  $0.3642$    &   $ 1.1490$       & $45.65$      & $46.72$ \\
    3D-FRONT &   $\mathbf{0.3371}$    & $\mathbf{1.1049}$       &  $\mathbf{50.01}$     & $\mathbf{52.91}$ \\
    \hline
    \end{tabular}
    \caption{\textbf{Evaluting diversity of scenes synthesized by models trained on 3D-FRONT vs.~SUNCG.}}
    \label{tab:diversity}%
\vspace{-4mm}
\end{table}%

\rz{In addition, we conduct a user study on AMT where Turkers were asked to choose scenes synthesized by~\cite{wang2018deep}, from randomly choosen empty rooms, that are deemed to be more ``plausible'';
see supplemental materials for details. From the user feedback, we find that layouts synthesized by the model trained on 3D-FRONT were chosen $64.8\%$ of the time (vs.~$35.2\%$ for SUNCG). All these results strongly demonstrate the utility of our new dataset, over other alternatives, for the important scene synthesis task.}


\subsection{Texturing 3D Models in Indoor Scenes}

\rz{
The quality, richness, and compatibility of object textures in an indoor scene can greatly enhance its realism. The textured 3D models
available from 3D-FRONT fulfill these very characteristics, and we expect our dataset to benefit the development of many data-driven 
scene texture synthesis algorithms. In comparison to texturing a single 3D object \cite{raj2019learning,gao2020tm,martinbrualla2020gelato}, doing the same to an object in the context of an indoor scene must
take into account that scene context to ensure both quality and visual compatibility.}


\rz{We extend a recent generative model for textured {\em meshes\/},} TM-Net \cite{gao2020tm}, to the 3D scene texturing task. TM-Net represents 3D shape parts with their structural deformable boxes, thus enables to generate part-level structural texture atlases for the given untextured 3D shapes. When applying it to the 3D scene configuration, we enforce the texture coherence between \rz{3D {\em objects\/}} by randomly choosing a shape in the scene, extracting its texture's VGG feature, and finally using the feature to guide the generation of other objects' textures in the training setting. After training the generative models, we synthesize texture for a random shape, and use it as a condition for other objects' texture generation to keep the consistency.

\begin{figure}[t]
\centering
\includegraphics[width = \linewidth]{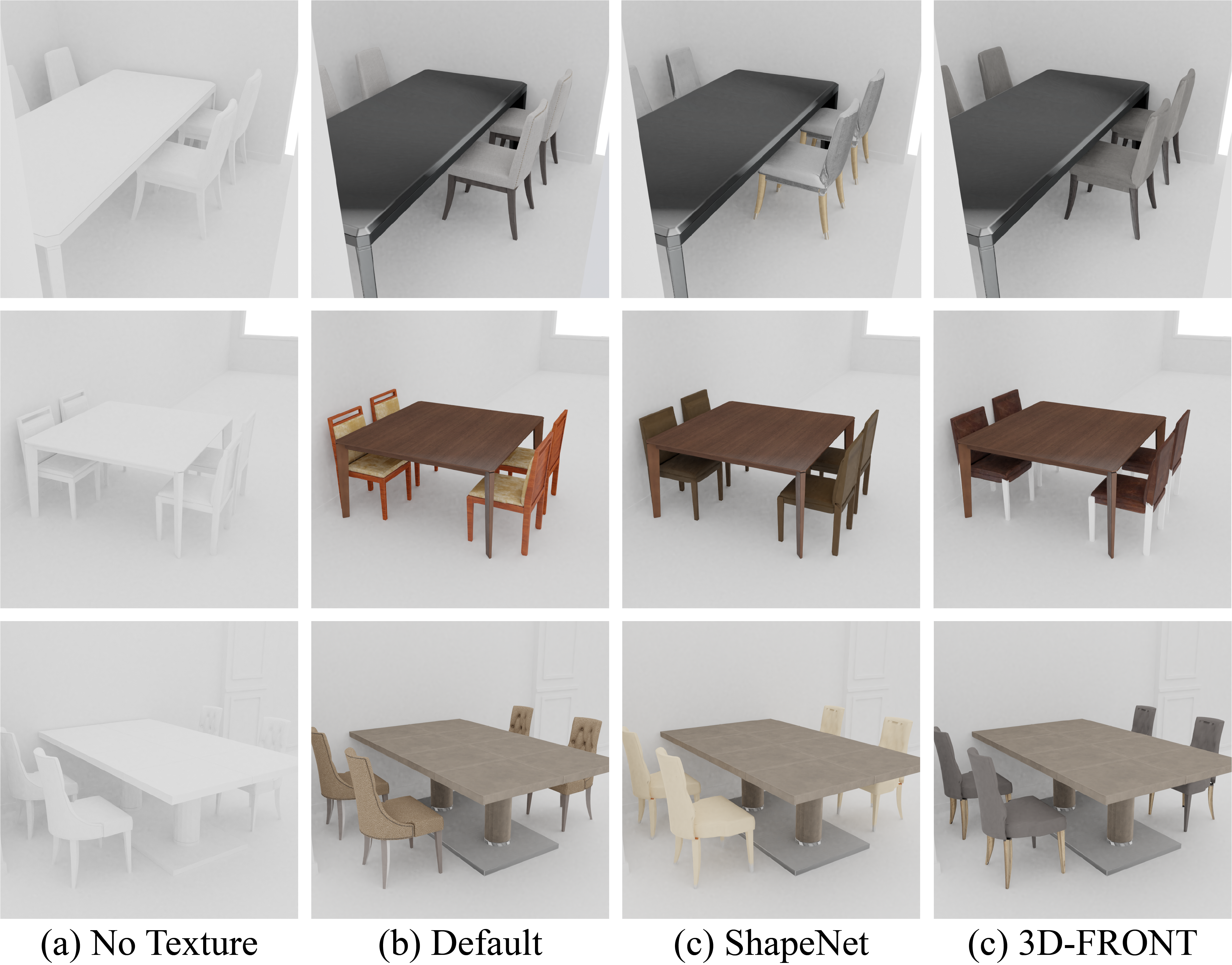}
\caption{\textbf{Texturing 3D models in indoor scenes.} 
\rz{The default textures (b) were provided by the 3D-FRONT dataset. In (c) and (d), we show chair textures generated by TM-Net, conditioned on given textures for the table. The network was trained on ShapeNet (c) or 3D-FRONT (d).}
}
\vspace{-4mm}
\label{fig:scene_tex}
\end{figure}

\rz{
We conduct a simple experiment to validate the advantage of 3D-FRONT, as training data for TM-Net, for the generation of chair textures given
table textures as a condition or guidance. Comparisons are made to ShapeNet \cite{chang2015shapenet}, which also contains textured 3D models and can serve as the training data. Figure~\ref{fig:scene_tex} presents some qualitative results where the table-chair settings were sampled from dining rooms. TM-Net trained on 3D-FRONT tends to generate richer and more diverse textures, as can be verified by both a quantitative test and a user study.
Specifically, the model trained on 3D-FRONT yields a LPIPS \cite{zhang2018perceptual} score of 0.289, which outperforms its ShapeNet counterpart, which has a score of 0.215, where we recall that LPIPS is a measure of the diversity of generated textures. 
Our user study on AMT, where users were asked to select which generated textures were ``richer'', also shows that results by TM-Net trained on 3D-FRONT were selected $61.1\%$ of the time (vs.~$38.9\%$ for ShapeNet); see supplemental material for more details.
}

\section{Conclusion and future work}
\label{sec:future}

We present 3D-FRONT, a new large-scale dataset of synthetic 3D indoor scenes. Up to now, there have been a variety of 3D scene datasets established to serve different purposes. Some focus on photorealistic renderings of artist-created scenes, possibly with instance segmentations 
and per-pixel material and illumination ground truth data, while others acquire large volumes of raw scans of the world to drive research in 3D scene reconstruction and modeling. Compared to these efforts, 3D-FRONT offers the largest publicly available collection of professional designed room layouts instanced with high-quality textured CAD meshes. 

One of our intentions was to fill a void in the vision and graphics community after SUNCG became unavailable. Yet, our dataset surpasses SUNCG in three aspects: professional vs.~amateur layout designs, CAD model quality, and style compatibility. We demonstrate that these distinctive features enable several data-driven applications which were not well supported by other datasets. 
In the future, we will continuously improve 3D-FRONT by releasing an industrial render engine (AceRay) and providing much enriched texture and 3D geometry contents.

\section*{Acknowledgement}
\label{sec:future}
We gratefully acknowledge the support of Alibaba Topping Homestyler with the sharing of the professional design templates and the high-quality 3D CAD models. We greatly thank the rendering service provided by the 3D rendering platform team at Tao Technology Department leading by Qian Qian. The released Treescope render is developed by the 3D scene platform team at Tao Technology Department leading by Xiaohang Hu.

{\small
\bibliographystyle{ieee_fullname}
\bibliography{egbib}
}

\section{Metrics in ``Validation and Assessment"}
\begin{figure}[t]
			\centering
			\includegraphics[width=\linewidth]{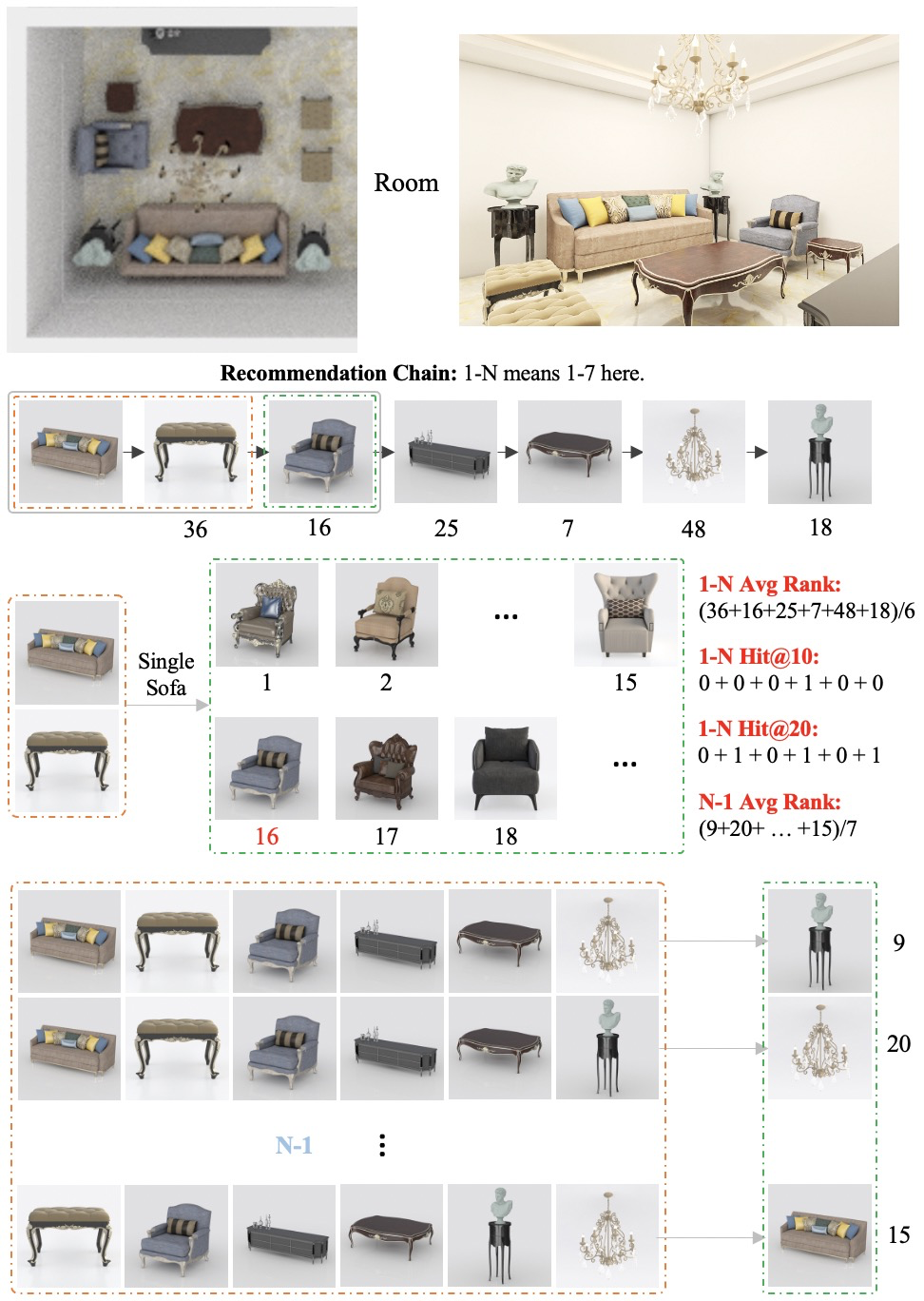}
			\caption{
			{\textbf{Metrics in ``Validation and Assessment".}} ``1-N" here means 1-7 because there are seven required objects in the room. When recommending a single sofa, the designer selected the \emph{16th} sofa from the retrieval sequence. Thus, the single sofa ranks 16 in this recommendation step. Hit@K refers to TopK recall accuracy \cite{sun2018pix3d}.
			}
			\label{sfig:metrics}
\end{figure}

We have briefly explained some metrics in Sec. 4 (Validation and Assessment) in the main paper. Here, we present an example in Figure~\ref{sfig:metrics} to make them more clear. 

\section{Other Statistics}

\begin{figure}[t]
			\centering
			\includegraphics[width=\linewidth]{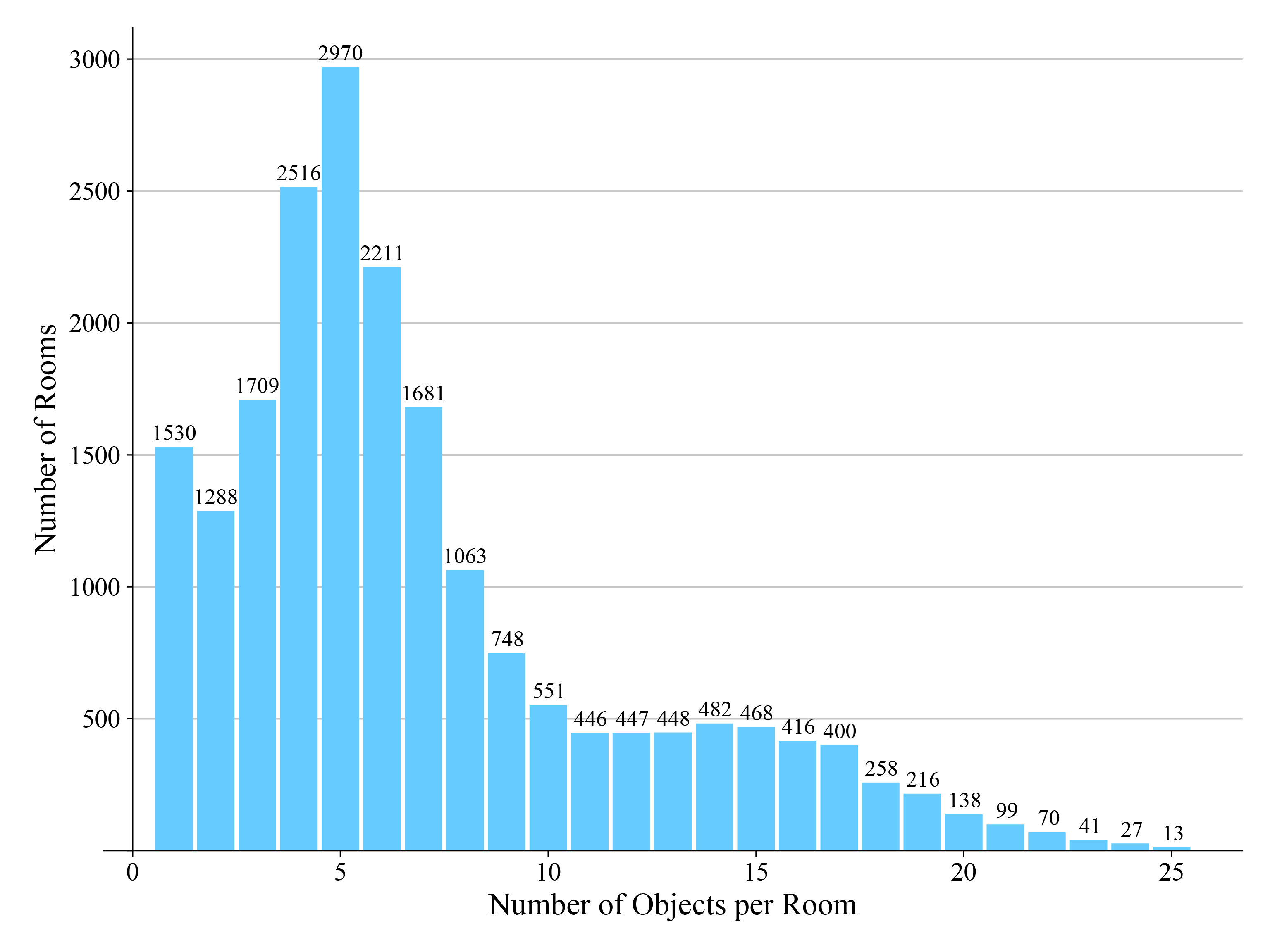}
			\caption{\LX{\textbf{Distribution of number of objects per room.}}
			}
			\label{sfig:model_room_num}
\end{figure}

\LX{In Figure~\ref{sfig:model_room_num}, we show the distribution of number of objects per room. At this time, 3D-FRONT's rooms are furnished by the functional furniture.}

\begin{figure*}[t]
			\centering
			\includegraphics[width=\linewidth]{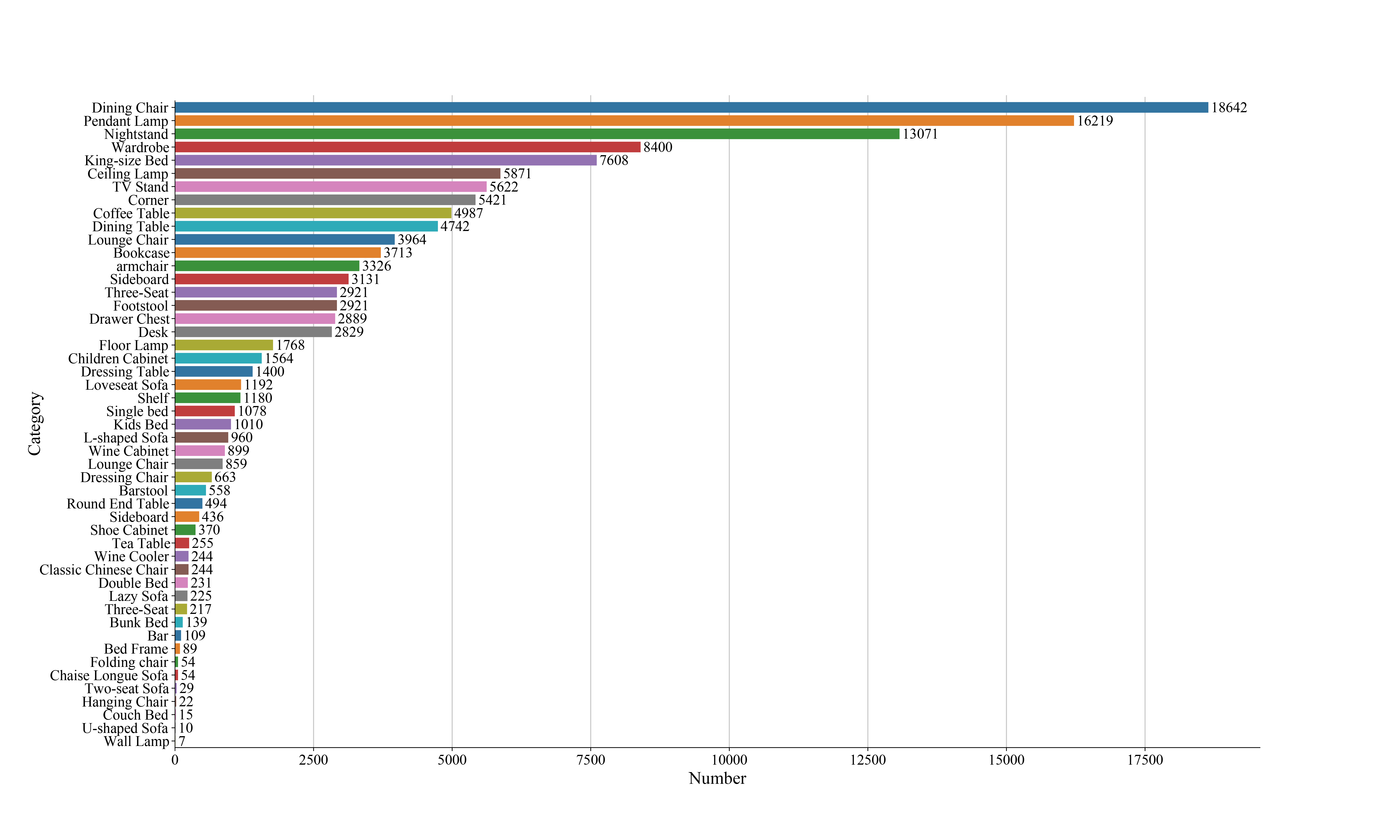}
			\caption{\textbf{Distribution of annotated instances} for 3D-FRONT's scenes corresponded to 3D-FUTURE's model categories.
			}
			\label{sfig:categorise}
\end{figure*}

In Figure~\ref{sfig:categorise}, we report the distribution of annotated object labels for 3D-FRONT's scenes corresponded to the 3D-FUTURE \cite{fu20203dfuture} 3D CAD model categories.

\begin{figure*}[t]
			\centering
			\includegraphics[width=\linewidth]{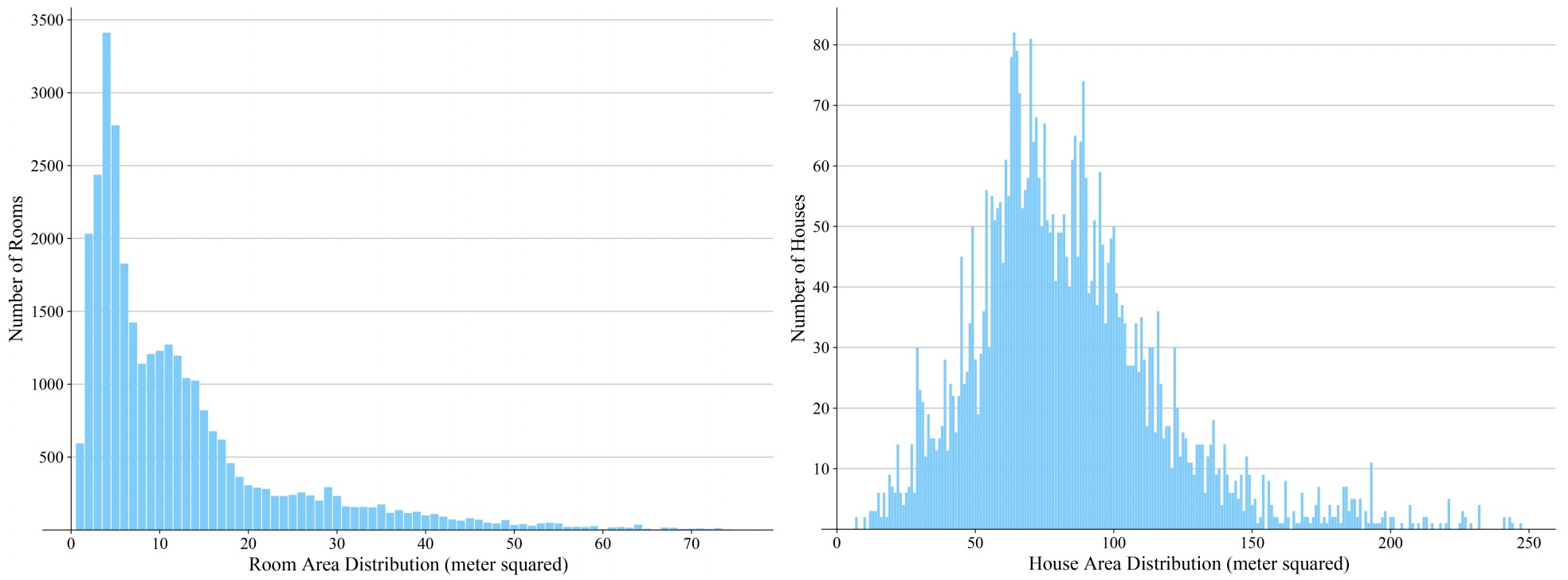}
			\caption{\textbf{Distribution of physical sizes} (in meters$^2$) per room (Left) and house (Right) of the 3D-FRONT dataset.
			}
			\label{sfig:model_room_cat}
\end{figure*}

\LX{In Figure~\ref{sfig:model_room_cat}, we show the distribution of object categories conditioned on different room types. The area of the square denotes the frequency of a given object category that appears in a certain room type. The frequency is normalized for each object category. It strongly implies the relationships between objects and rooms. For example, categories such as children cabinet, bunk bed, and kids bed are more likely to appear in a kid room, while bookcase and desk are more likely to appear in a study room. We can learn rich design knowledge from the distribution.}

\begin{figure*}[t]
			\centering
			\includegraphics[width=0.95\linewidth]{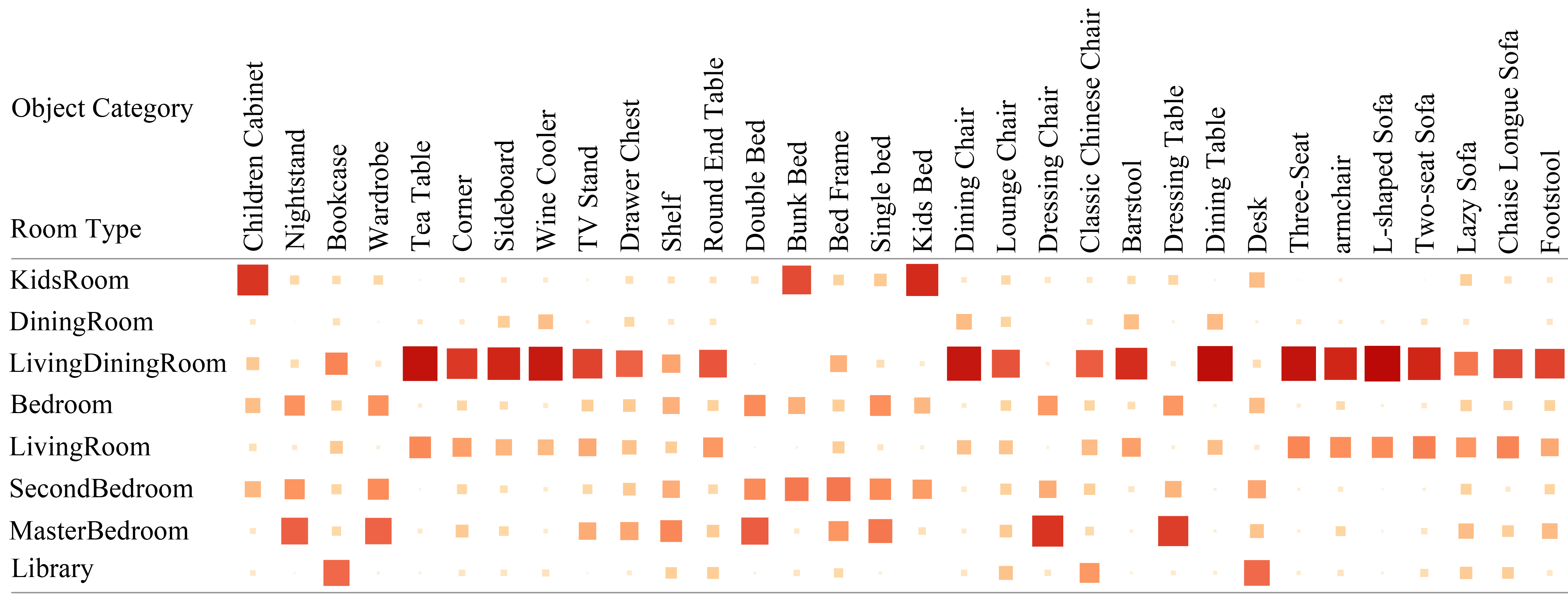}
			\caption{\LX{\textbf{Distribution of object categories conditioned on different room types.}}
			}
			\label{sfig:area_size}
\end{figure*}

\LX{In Figure~\ref{sfig:area_size}, we present the physical sizes over the rooms and houses. The 3D-FRONT dataset are measured in real-world spatial dimensions (units are in meters).}

\begin{figure*}[t]
			\centering
			\includegraphics[width=\linewidth]{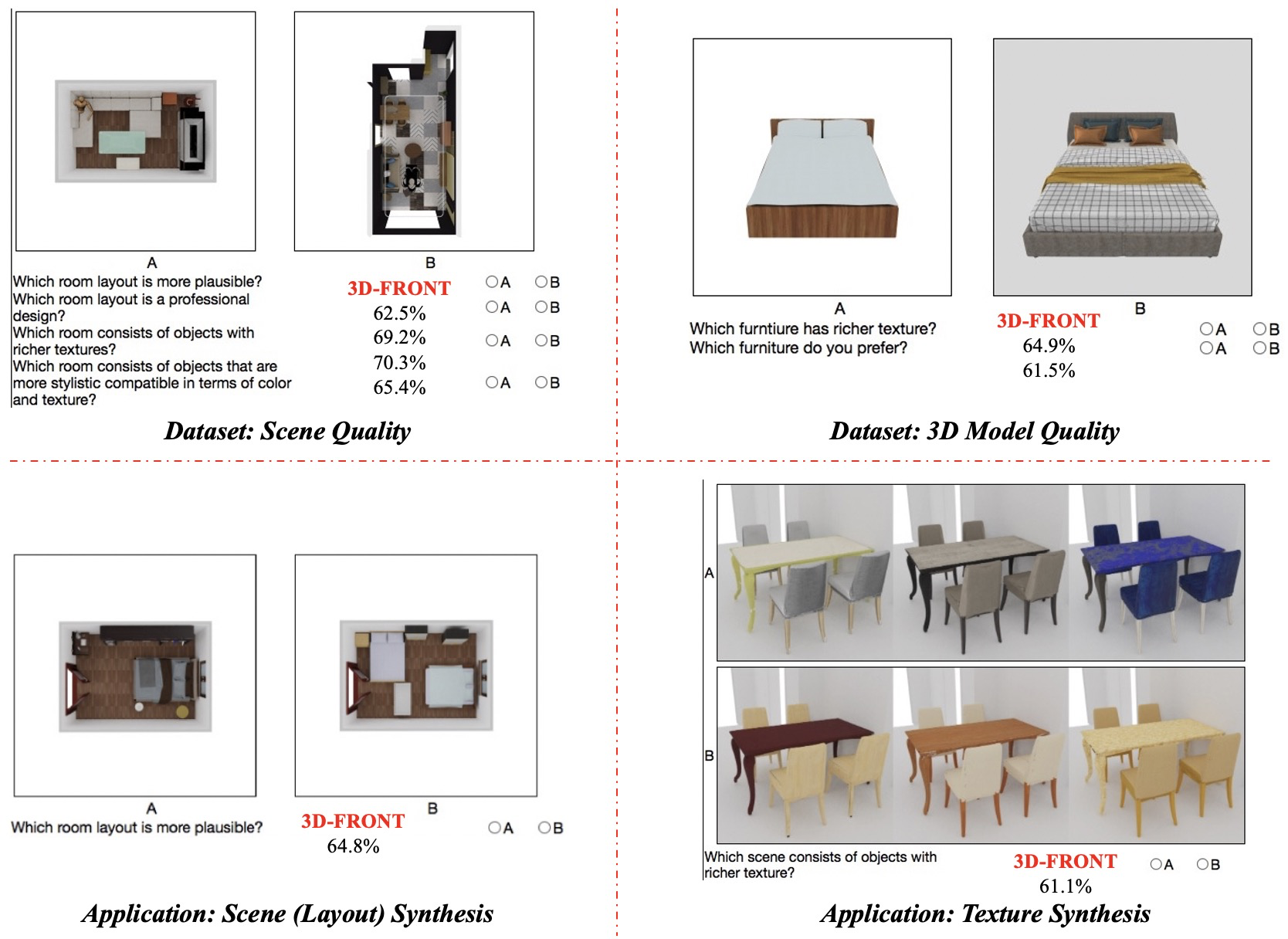}
			\caption{\textbf{User Studies.} We perform the listed user studies using Amazon Mechanical Turk (AMT). 3D-FRONT's scores are reported in the figure. From the dataset comparisons, we see that for each quality criterion assessed, the majority of Turkers (between 60\% and 70\%), preferred data presented by 3D-FRONT. See Sec.~\ref{sec:us} for the experimental settings.
			}
			\label{fig:user-study}
\end{figure*}

\HF{\section{User Studies}}
\label{sec:us}
We perform user studies to show the quality of 3D-FRONT using Amazon Mechanical Turk (AMT). The compared datasets are 3D-FRONT, SUNCG \cite{song2016ssc}, and ShapeNet \cite{chang2015shapenet}. The questions and the scores are shown in Figure~\ref{fig:user-study}. The tasks are explained below.
\newline

\noindent \textbf{Dataset: \emph{Scene Quality}.} We have 90 pairs of scenes randomly sampled from SUNCG and 3D-FRONT based on scene types (LivingRoom, DiningRoom, and Bed Room) in our questionnaire. Each scene type contains 30 pairs. We study layout plausibility, design quality, texture quality, and style compatibility in this task. We have collected 20 questionnaires from master-level annotators in AMT. That means each scene pair has been labeled by 20 annotators. Thus, the final scores are calculated using 1,800 feedback.
\newline

\noindent \textbf{Dataset: \emph{Model Quality}.} We have 30 pairs of furniture models randomly sampled from SUNCG and 3D-FRONT based on categories in our questionnaire. We study texture quality and model's visual quality in this task. We have collected 20 questionnaires from master-level annotators in AMT. Thus, the final scores are calculated using 600 feedback.

\noindent \textbf{Application: \emph{Layout Synthesis}.} We randomly sampled 60 rooms from 3D-FRONT (36) and SUNCG (24). For each room, we synthesizing a pair of layouts, with one produced by the model trained on 3D-FRONT and the other generated by the model trained on SUNCG. We study layout plausibility in this task. We have collected 20 questionnaires from master-level annotators in AMT. Thus, the final score is calculated using 1,200 feedback.

\noindent \textbf{Application: \emph{Texture Synthesis}.} We have selected 5 DiningRoom corners from 3D-FRONT and textured their chairs and tables using the learned texture synthesis models (3D-FRONT vs. ShapeNet). For each corner, we have perform texture synthesis three times with random noises. We study texture diversity in this task. We have collected 20 questionnaires from master-level annotators in AMT. Thus, the final score is calculated using 100 feedback.
\newline

\section{More House \& Room Examples}
\begin{figure*}[t]
			\centering
			\includegraphics[width=\linewidth]{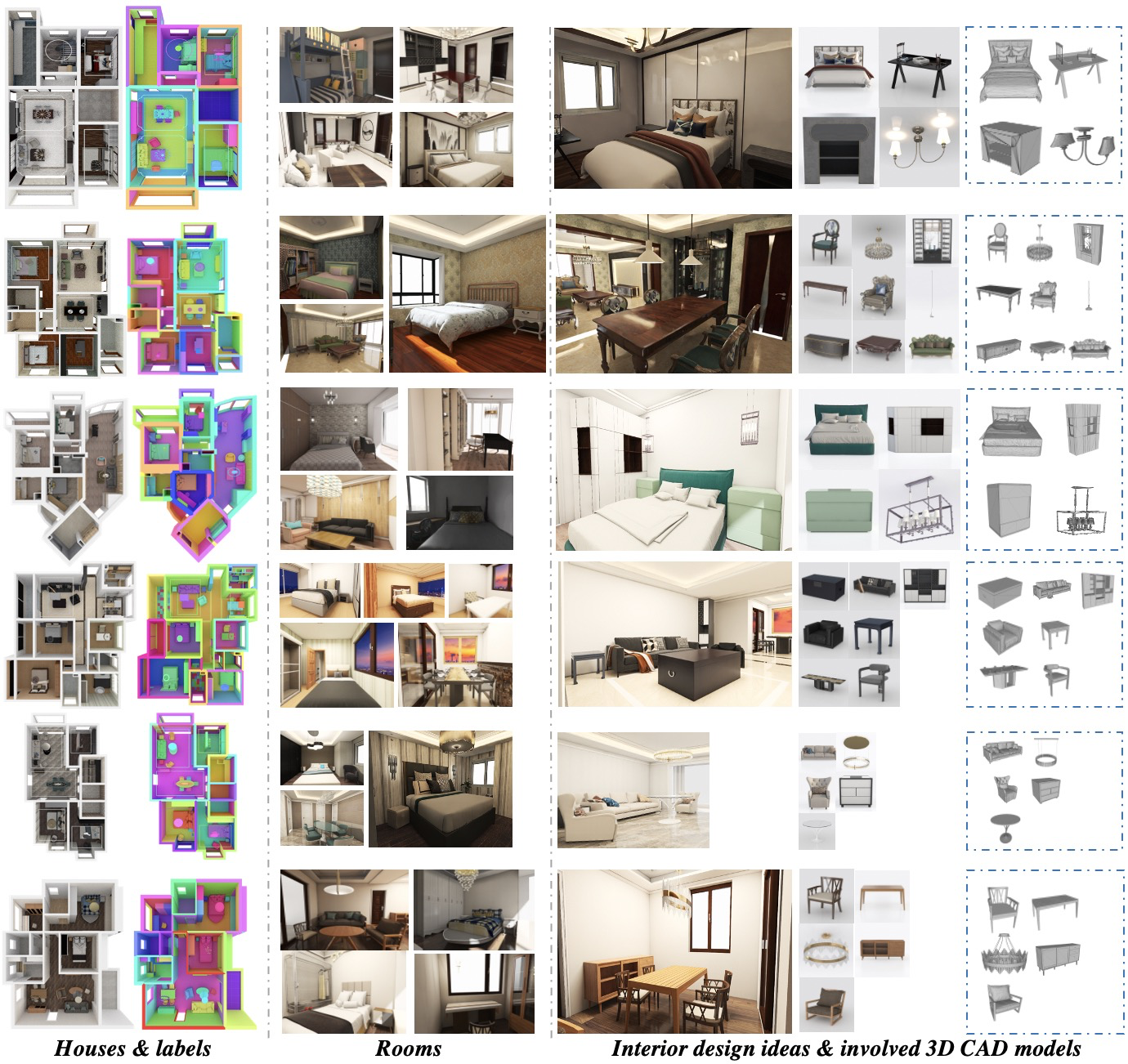}
			\caption{
			{\textbf{House Examples - Part 1.}} Zoom in for better view. 
			}
			\label{sfig:example1}
\end{figure*}

\begin{figure*}[t]
			\centering
			\includegraphics[width=\linewidth]{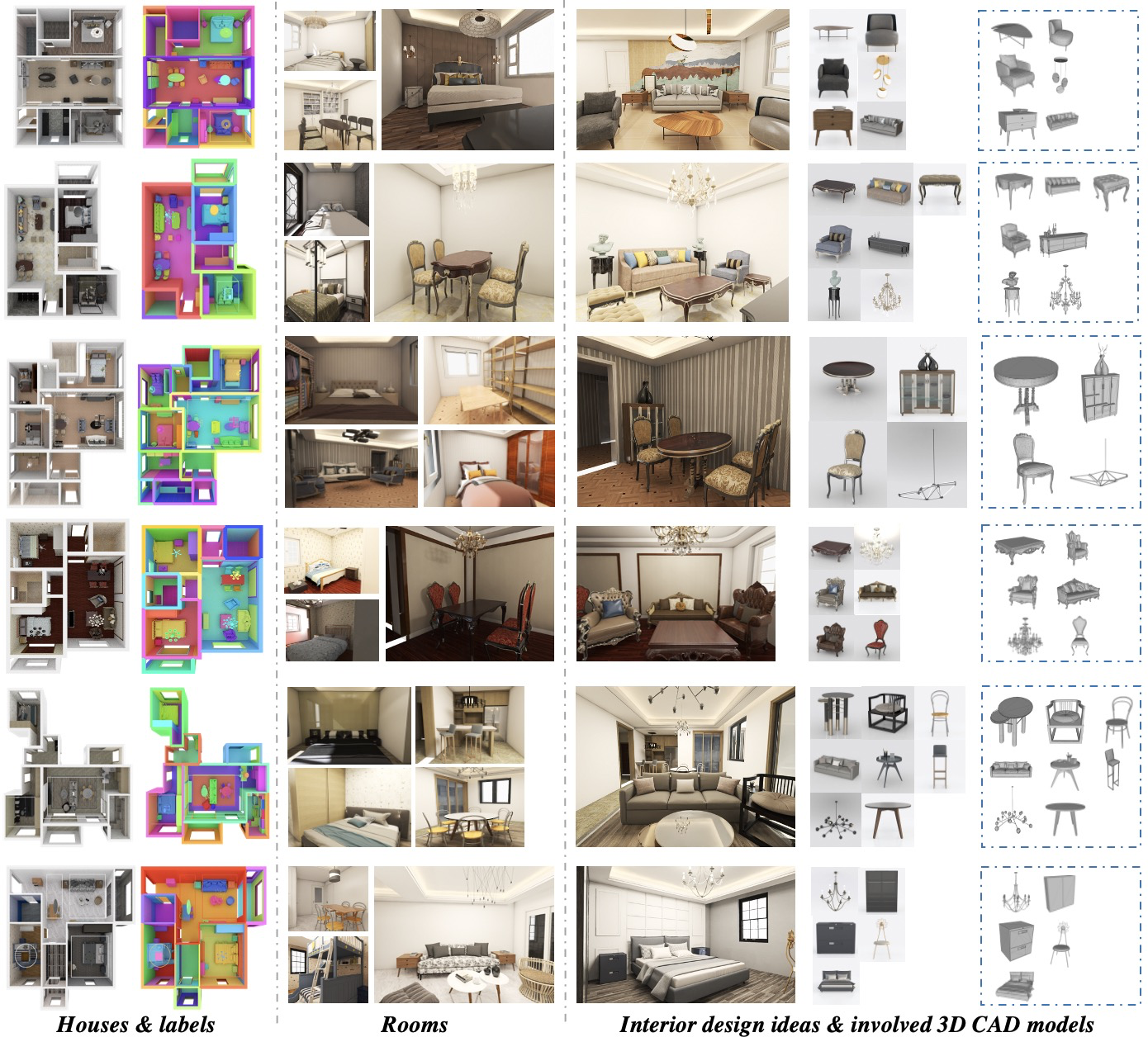}
			\caption{
			{\textbf{House Examples - Part 2.}} Zoom in for better view. 
			}
			\label{sfig:example2}
\end{figure*}

\begin{figure*}[t]
			\centering
			\includegraphics[width=\linewidth]{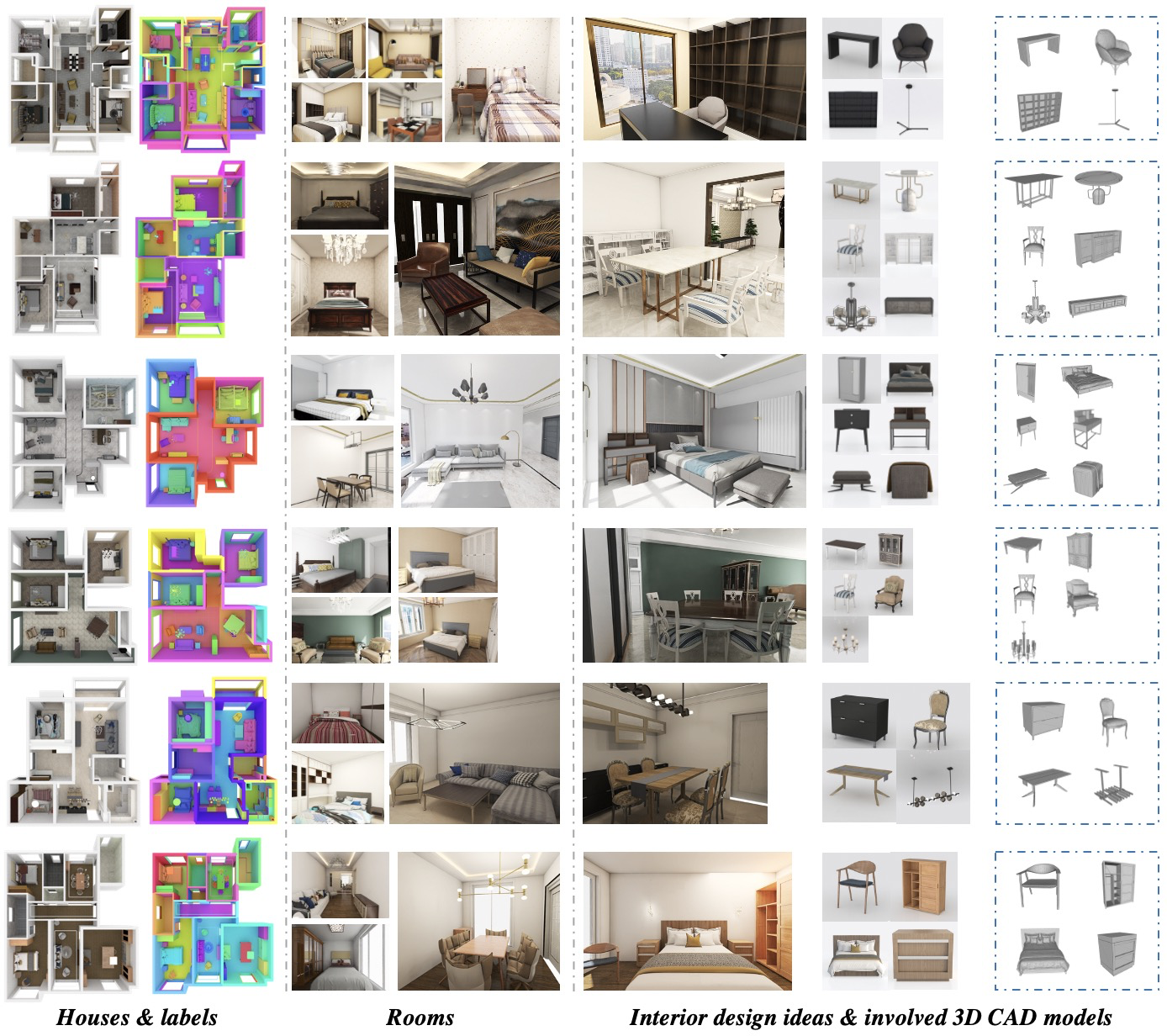}
			\caption{
			{\textbf{House Examples - Part 3.}} Zoom in for better view. 
			}
			\label{sfig:example3}
\end{figure*}

\begin{figure*}[t]
			\centering
			\includegraphics[width=\linewidth]{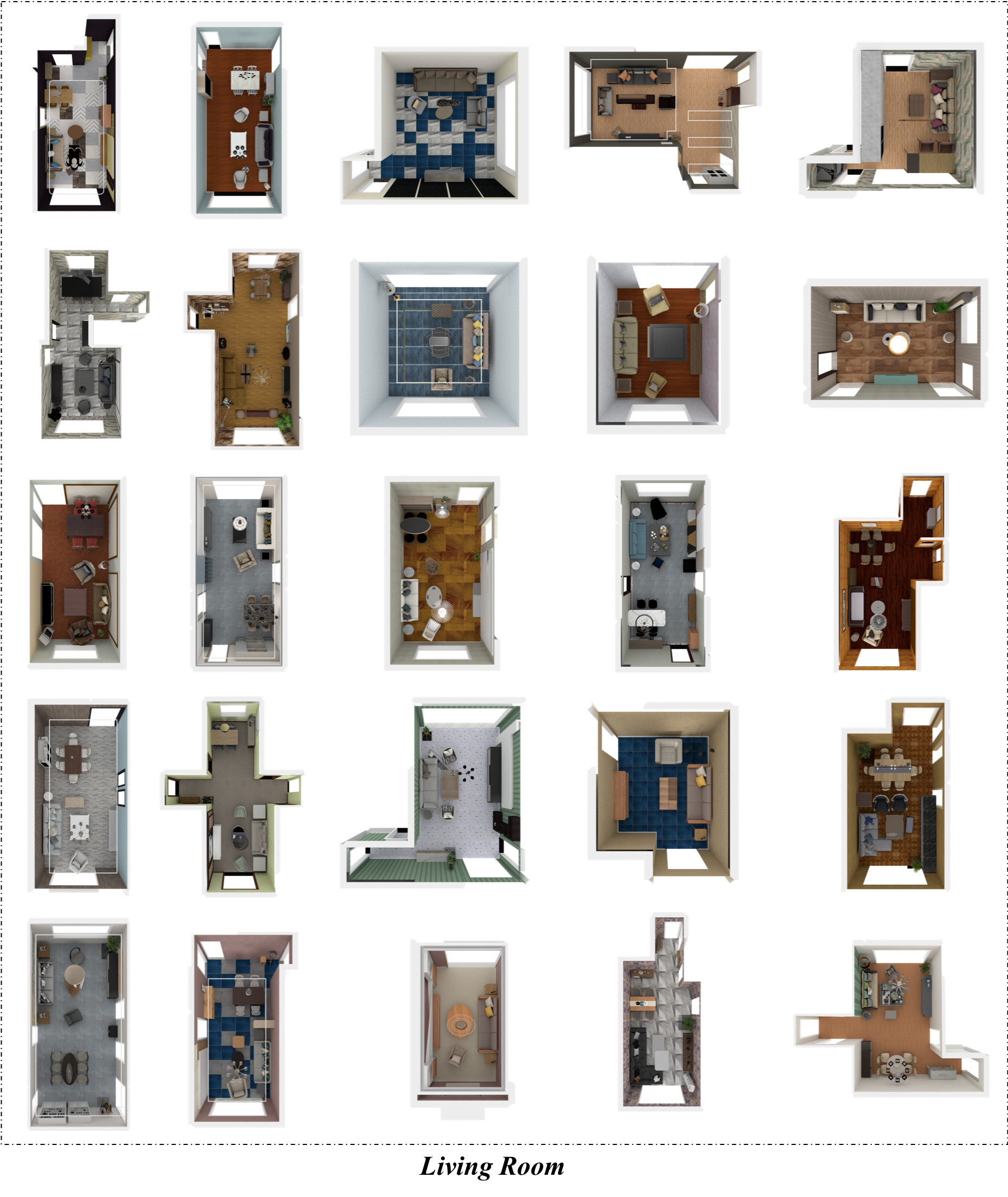}
			\caption{
			{\textbf{Room Examples - Part 1.}} Zoom in for better view. 
			}
			\label{sfig:room_example1}
\end{figure*}

\begin{figure*}[t]
			\centering
			\includegraphics[width=\linewidth]{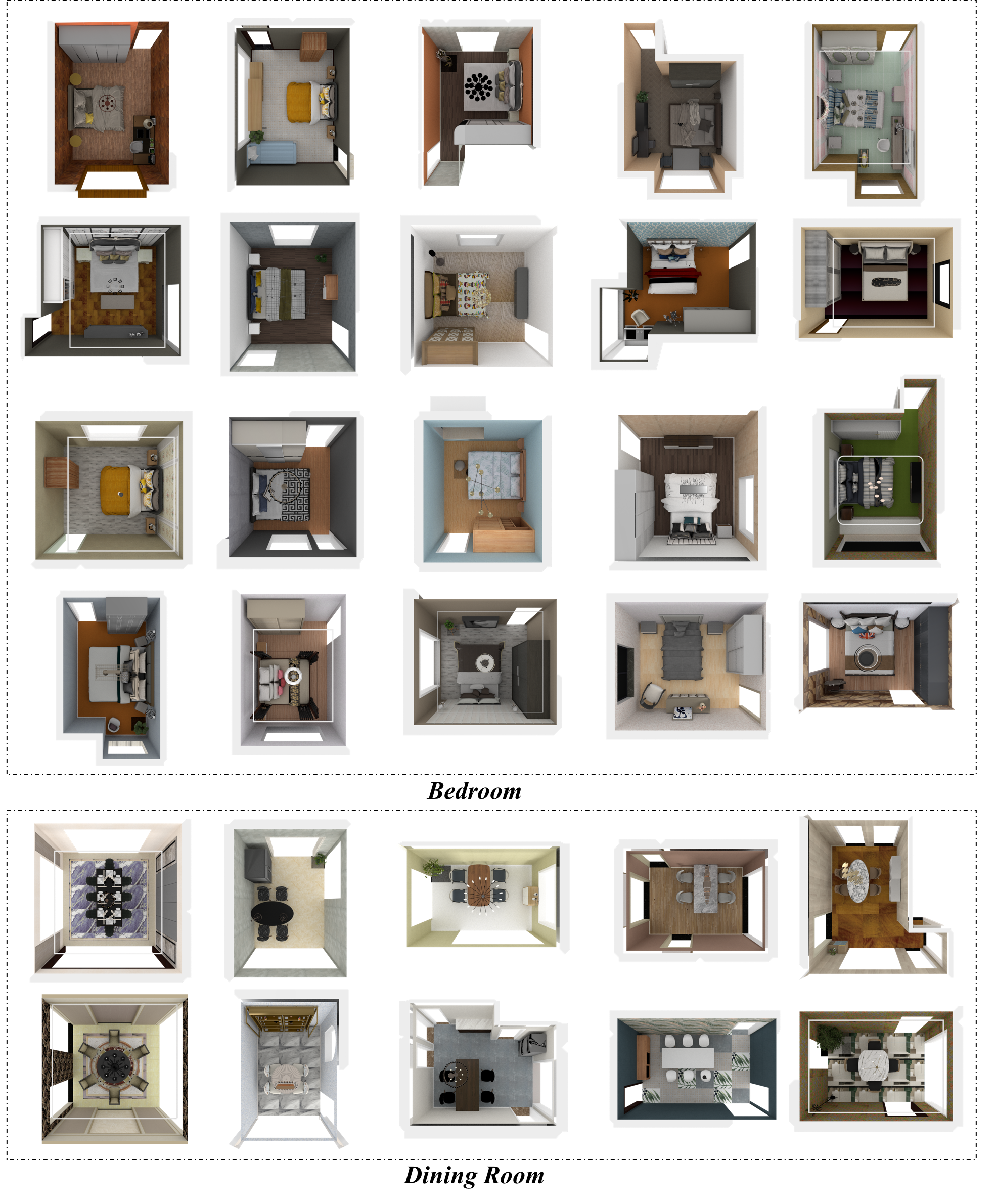}
			\caption{
			{\textbf{Room Examples - Part 2.}} Zoom in for better view. 
			}
			\label{sfig:room_example2}
\end{figure*}

In Figure~\ref{sfig:example1}, Figure~\ref{sfig:example2}, Figure~\ref{sfig:example3}, Figure~\ref{sfig:room_example1}, and Figure~\ref{sfig:room_example2}, we present more house and room examples to demonstrate the quality of 3D-FRONT.

\end{document}